\documentclass[conference]{IEEEtran}
\usepackage{times}

\usepackage[numbers]{natbib}
\usepackage{multicol}
\usepackage{multirow}
\usepackage[bookmarks=true]{hyperref}

\usepackage{booktabs}
\usepackage{afterpage}
\usepackage{graphics} 
\usepackage{graphicx}
\usepackage{array}
\usepackage{relsize}
\usepackage{multirow}
\usepackage{amsmath}
\usepackage{amsfonts, bm}
\usepackage{amsthm}
\usepackage{amssymb}
\usepackage{algorithm}
\usepackage{algorithmic}
\usepackage{wrapfig}
\usepackage{enumerate}
\usepackage{color}
\usepackage{subfigure}
\usepackage{xspace}
\usepackage[normalem]{ulem}
\usepackage{tikz}
\usepackage{rotating}
\usepackage{bbm}

\usepackage[small,it]{caption}

\definecolor{newTxtColor}{rgb}{0 0.45 0}
\definecolor{chopColor}{rgb}{0.45 0 0}

\newcommand{\tabTopBotLnWd}{1pt}
\newcommand{\todo}[1]{\textcolor{blue}{\textbf{[#1]}}}
\newcommand{\newTxt}[1]{#1}

\newcommand{\header}[1]{\smallskip\noindent\textbf{#1} }

\newcommand{\Lzero}{{$L_0$}}
\newcommand{\Lone}{{$L_1$}}
\newcommand{\Ltwo}{{$L_2$}}
\newcommand{\RGBD}{\mbox{RGB-D }}

\begin{document}

\title{Deep Learning for Detecting Robotic Grasps \vspace*{0in}}


\author{
Ian Lenz,$^\dagger$ Honglak Lee,$^*$ and Ashutosh Saxena$^\dagger$\\
$^\dagger$ Department of Computer Science, Cornell University.\\
$^*$ Department of EECS, University of Michigan, Ann Arbor.\\
Email: ianlenz@cs.cornell.edu, honglak@eecs.umich.edu, asaxena@cs.cornell.edu
}

\maketitle

\IEEEpeerreviewmaketitle
\vspace*{0in}



\newenvironment{packed_enum}{
\begin{enumerate}
  \setlength{\itemsep}{0pt}
  \setlength{\parskip}{0pt}
  \setlength{\parsep}{0pt}
}
{\end{enumerate}}

\newenvironment{packed_item}{
\begin{itemize}
  \setlength{\itemsep}{0pt}
  \setlength{\parskip}{0pt}
  \setlength{\parsep}{0pt}
}{\end{itemize}}

\newlength\savedwidth
\newcommand\whline[1]{\noalign{\global\savedwidth\arrayrulewidth
                               \global\arrayrulewidth #1} %
                      \hline
                      \noalign{\global\arrayrulewidth\savedwidth}}
\renewcommand\multirowsetup{\centering}

\newlength{\sectionReduceTop}
\newlength{\sectionReduceBot}
\newlength{\subsectionReduceTop}
\newlength{\subsectionReduceBot}
\newlength{\abstractReduceTop}
\newlength{\abstractReduceBot}
\newlength{\captionReduceTop}
\newlength{\captionReduceBot}
\newlength{\subsubsectionReduceTop}
\newlength{\subsubsectionReduceBot}
\newlength{\headerReduceTop}
\newlength{\figureReduceBot}

\newlength{\horSkip}
\newlength{\verSkip}

\newlength{\equationReduceTop}

\newlength{\figureHeight}
\setlength{\figureHeight}{1.7in}

\setlength{\horSkip}{-.09in}
\setlength{\verSkip}{-.1in}

\setlength{\figureReduceBot}{-0.0in}
\setlength{\headerReduceTop}{0.05in}
\setlength{\subsectionReduceTop}{-0.0in}
\setlength{\subsectionReduceBot}{-0.0in}
\setlength{\sectionReduceTop}{-0.0in}
\setlength{\sectionReduceBot}{-0.0in}
\setlength{\subsubsectionReduceTop}{-0.0in}
\setlength{\subsubsectionReduceBot}{-0.0in}
\setlength{\abstractReduceTop}{-0.0in}
\setlength{\abstractReduceBot}{-0.0in}

\setlength{\equationReduceTop}{-0in}

\setlength{\captionReduceTop}{-0.0in}
\setlength{\captionReduceBot}{-0.03in}

%

\newcommand{\fix}{\marginpar{FIX}}
\newcommand{\new}{\marginpar{NEW}}


\maketitle



\begin{abstract}
We consider the problem of detecting robotic grasps in
an \RGBD view of a scene containing objects. 
In this work, we apply a deep learning approach to solve this problem,
which avoids time-consuming hand-design of features.
  This presents two main challenges.
First, we need to evaluate a huge number of candidate grasps.
In order to make detection fast and robust, we present a two-step cascaded system with two deep networks, where the top detections from the first are re-evaluated by the second. 
The first network has fewer features, is faster to run, and can effectively prune out unlikely candidate grasps. 
The second, with more features, is slower but has to run only on the top few
detections.
Second, we need to handle multimodal inputs effectively, for which we present a method that applies structured regularization
on the weights based on multimodal group regularization.
We show that our method improves performance on an RGBD robotic grasping dataset,
and can be used to successfully execute grasps on two different robotic platforms.
\footnote{Parts of this work were presented at ICLR 2013 
as a workshop paper, and at RSS 2013 as a conference paper. This version
includes significantly extended related work, algorithmic descriptions, and
extensive robotic experiments which were not present in previous versions.}

\end{abstract}

\medskip


\section{Introduction}

\begin{figure*}[tb]
\begin{center}
\includegraphics[height=1.75in]{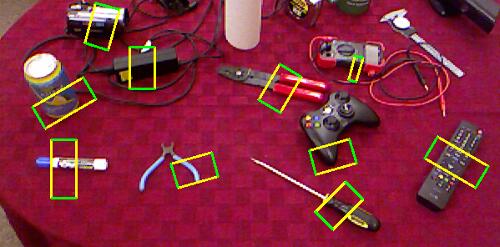} 
\hskip .2in
\includegraphics[height=1.75in]{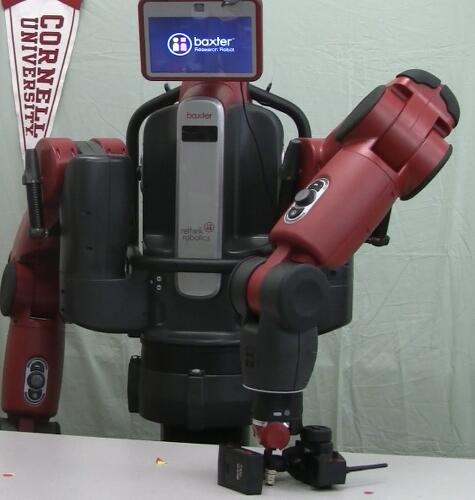} \hspace*{0.2in}
\begin{minipage}{0.15\textwidth}
\vspace*{-1.7in}
\includegraphics[width=\textwidth]{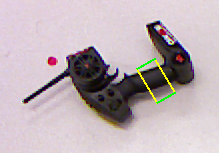}
\vspace*{0.1in} \\
\includegraphics[width=\textwidth]{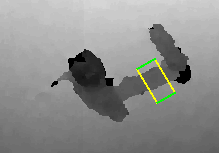}
\end{minipage}
\end{center}
\vspace*{\captionReduceTop}
\caption{\textbf{Detecting robotic grasps:} Left: A cluttered lab scene labeled
with rectangles corresponding to robotic grasps for objects in the scene.
Green lines correspond to robotic gripper plates. We use a two-stage system
based on deep learning to learn features and perform detection for robotic
grasping. Center: Our Baxter robot ``Yogi'' successfully executing a 
grasp detected by our algorithm. Right: The grasp detected for this case,
in the RGB (top) and depth (bottom) images obtained from Kinect.}
\label{fig:gripper}
\vspace*{\captionReduceBot}
\vspace*{\figureReduceBot}
\end{figure*}

Robotic grasping is a challenging problem involving perception, planning, and control. Some recent works \cite{saxena2006roboticgrasping,SaxenaGraspingAAAI,JiangICRA2011,HsiaoKaijenGraspingPaper}
address the perception aspect of this problem by converting it
into a detection problem
in which, given a noisy, partial view of the object from a 
camera, the goal
is to infer the top locations where a robotic gripper could be placed (see Figure~\ref{fig:gripper}).
Unlike generic vision problems based on static images, such robotic perception problems
are often used in closed loop with controllers, so there are stringent
requirements on performance and computational speed. 
In the past, hand-designing features has been the most popular method
for several robotic tasks \cite{MaitinTowel,Kragic03robustvisual}.
However, this is cumbersome and time-consuming, especially when
we must incorporate new input modalities such as \RGBD cameras.


Recent methods based on deep learning~\cite{bengio2009learning} have demonstrated state-of-the-art performance in a wide
variety of tasks, including visual recognition~\cite{quocLargeScale,Sohn+etal:iccv2011}, 
audio recognition~\cite{convDBNAudio,mohamed2012acoustic}, 
and natural language processing \cite{NLPfromScratch}. These techniques
are especially powerful because they are capable of learning useful features
directly from both unlabeled and labeled data, avoiding the need for hand-engineering.

However, most work in deep learning has been applied in the context of
\emph{recognition}.
Grasping is inherently a \emph{detection} problem,
and previous applications of deep learning to detection have typically focused on specific vision applications such as face detection \cite{osadchy2007synergistic} and pedestrian detection~\cite{sermanet-cvpr-2013}. 
Our goal is not only to infer a viable grasp, but to infer the optimal grasp for a given object that maximizes the chance of successfully grasping it, which 
differs significantly from the problem of object detection. 
Thus, the first major contribution of our work is to apply deep learning
to the problem of robotic grasping, in a fashion which could generalize
to similar detection problems.

The second major contribution of our work is to propose a new method for
handling multimodal data in the context of feature learning. 
The use of
\RGBD data, as opposed to simple 2D image data, has been shown to
significantly improve grasp detection results \cite{JiangICRA2011,Dogar-RSS-12,SaxenaGraspingAAAI}.
In this work, we present a multimodal feature learning algorithm which adds a 
structured regularization penalty to the objective function to be optimized
during learning. 
%
%
As opposed to previous works in deep
learning, which either ignore modality information at the first layer (i.e., encourage all features to use all modalities) \cite{SocherNIPS2012} or
train separate first-layer features for each modality \cite{multimodal,srivastava2012multimodal}, our
approach allows for a middle-ground in which each feature is encouraged to
use only a subset of the input modalities, but is not forced to use only
particular ones.

We also propose a two-stage cascaded detection system based on deep 
learning. Here, we use fewer features for the first pass, providing
faster, but only approximately accurate detections. The second pass uses
more features, giving more accurate detections.  
In our experiments, we found that the first deep network, with fewer
features, was better at avoiding overfitting but less accurate.  
We feed the top-ranked rectangles from the first layer into
the second layer, leading to robust early rejection of false positives.
Unlike manually designed two-step features as in \cite{JiangICRA2011},
our method uses deep learning, which allows us to learn detectors that not only
give higher performance, but are also computationally efficient. 


We test our approach on a challenging dataset, where 
we show that our algorithm improves both recognition and detection performance 
for grasping rectangle data. We also show that our two-stage approach 
is not only able to match the performance of a single-stage 
system, but, in fact, improves results while
significantly reducing the computational time needed for detection.

In summary, the contributions of this paper are:
\begin{packed_item}
\item We present a deep learning algorithm for detecting robotic grasps. To the best of our knowledge, this is the first work to do so.
\item In order to handle multimodal inputs, we present a new way to apply
structured regularization to the weights to these inputs based 
on multimodal group regularization.
\item We present a
multi-step cascaded system for detection, significantly reducing its
computational cost. 
\item Our method outperforms the state-of-the-art for rectangle-based 
grasp detection,
as well as previous deep learning algorithms.
\item We implement our algorithm on both a Baxter and a PR2 robot, and
show success rates of 84\% and 89\%, respectively, for executing grasps on a
highly varied set of objects.
\end{packed_item}

The rest of the paper is organized as follows: We discuss
related work in Section~\ref{sec:relatedwork}. 
We present our two-step cascaded detection system in 
Section~\ref{sec:detection}, and some additional details in
Section~\ref{sec:input}. We then describe our feature learning algorithm
and structured regularization method in Section~\ref{sec:sparsity}.
We present our experiments in Section~\ref{sec:experiments}, and 
discuss results in Section~\ref{sec:results}. We then present experiments on
both Baxter and PR2 robots
in Section~\ref{sec:baxter}. We present several interesting directions for
future work in Section~\ref{sec:future}, then conclude in 
Section~\ref{sec:conclusion}.


\section{Related Work\label{sec:relatedwork}}



\subsection{Robotic Grasping}
\newTxt{
In this section, we will 
focus on perception- and learning-based approaches for robotic grasping. For a more complete
review of the field, we refer the reader to review papers by
\citet{bohgReview,SahbaniSurvey,BicchiSurvey} and \citet{ShimogaSurvey}.

Most works define a ``grasp'' as an end-effector configuration which achieves
partial or complete form- or force-closure of a given object. This is a
challenging problem because it depends on the pose and configuration of the
robotic gripper as well as the shape and physical properties of the object
to be grasped, and typically requires a search over a large number of
possible gripper configurations. Early works 
\cite{lakshminarayana1978mechanics,nguyen1986,ponce1993computing} 
focused on testing for 
form- and force-closure, and synthesizing grasps fulfilling these properties
according to some hand-designed ``quality score'' \cite{Ferrari1992}.
More recent works have refined these definitions
\cite{Rodriguez-RSS-11}.
These works assumed full knowledge of object shape and physical properties.

\header{Grasping Given 3D Model:}
Fast synthesis of grasps for known 3D models remains an active 
research topic \cite{Dogar-RSS-12,graspit1,WeiszICRA12}, 
with recent methods using advanced physical 
simulation to find optimal grasps. 
\citet{Gallegos-RSS-11} performed optimization of 
grasps given both a 3D model of
the object to be grasped and the desired contact points for the robotic
gripper. \citet{pokorny2013b} define spaces of graspable objects, then map new
objects to these spaces to discover grasps. However, these works are only 
applicable when the full 3D model of the object is exactly known, which may
not be the case when a robot is interacting with a new environment. We note that
some of these physics-based approaches might be combined with our approach in
a multi-pass system, discussed further in Sec.~\ref{sec:future}.

\header{Sensing for Grasping:}
In a real-world robotic setting, a robot will not have full knowledge of
the 3D model and pose of an object to be grasped, but rather only incomplete
information from some set of sensors such as color or depth cameras, tactile
sensors, etc.
This makes the problem of grasping significantly more challenging
\cite{bohgReview},
as the algorithm must use more limited and potentially noisier information
to detect a good grasp. While some works \cite{ColletRomeaICRA2009,
PapazovIJRR2012} simply attempt to estimate the poses of known objects and then
apply full-model grasping algorithms based on these results, others 
avoid this assumption, functioning on novel objects
which the algorithm has not seen before. 

Such works often made use of other simplifying assumptions, 
such as assuming that objects belong to one
of a set of primitive shapes \cite{piater2000,BowersLumia2003}, or are planar
\cite{MoralesIROS2002}. Other works
produced impressive results for specific cases, such as grasping the
corners of towels \cite{MaitinTowel}.
While such works escape the
assumption of a fully-known object model, hand-coded grasping rules have
a hard time dealing with the wide range of objects seen in real-world human
environments, and are difficult and time-consuming to create.

\header{Learning for Grasping:}
Machine learning methods have proven effective for a wide range of perception
problems 
\cite{ViolaJonesCascaded,HintonScience,convNet,SocherNIPS2012,bo_iser12},
allowing a perception system to learn a mapping from some
feature set to various visual properties. Early work by \citet{KamonICRA96}
showed that learning approaches could also be applied to the
problem of grasping from vision, introducing a learning component to grasp
quality scores.

Recent works have employed richer
features and learning methods, allowing robots to grasp 
known objects which might be partially occluded \cite{Glover-RSS08}
or in an unknown pose \cite{detryICDL2009} as well as fully novel objects
which the system has not seen before \cite{saxena2006roboticgrasping}. Here,
we will address the latter case. Earlier
work focused on detecting only a single grasping point 
from 2D partial-view data,
using heuristic methods to determine a gripper pose based on this point.
\cite{Saxena-IJRR-Grasping}. The use of 3D data was shown to 
significantly improve
these results \cite{SaxenaGraspingAAAI} thanks to giving direct physical
information about the object in question. With the advent of low-cost 
\RGBD sensors such as the Kinect, the use of depth data for robotic 
grasping has become ubiquitous.

Several other works attempted to use the learning algorithm to more fully
constrain the detected grasps. 
\citet{EkvallK07} and \citet{HuebnerK08} used shape-based approximations as
bases for learning algorithms which directly gave an approach
vector.
\citet{QuocLeRankToGrasp} treated grasp detection as a ranking problem
over sets of contact points in image space.   
\citet{JiangICRA2011} represented a grasp as a 2D oriented rectangle 
in image space, with two edges corresponding to the gripper plates, using
surface normals to determine the grasp approach vector. These
approaches allow the detection algorithm to detect more exactly the gripper
pose which should be used for grasping. In this work, we will follow the
rectangle-based method.


Learning-based approaches have shown impressive results in grasping novel
objects, showing that learning some parameters of the detection system
can outperform human tuning. However,
these approaches still require a significant degree of hand-engineering in
the form of designing good input features.

\header{Other Applications with RGBD Data.}
Due to the availability of inexpensive depth sensors, 
\RGBD data has been a significant research focus in
recent years for various robotics applications. For example, 
\citet{jiang2012placingobjects} 
consider robotic placement of objects, while \citet{Teuliere12a} used
\RGBD data for visual servoing.
Several works, including those of \citet{endres2013ijrr} and 
\citet{Whelan13icra} have
extended and improved Simultaneous Localization and Mapping (SLAM) 
for \RGBD data.
Object detection and recognition has been a major focus in research on
\RGBD data \cite{Collet_Romea_2011_6856,LaiICRA11,cadenaICRA13}. 
Most such works use hand-engineered features such as
\cite{VFH}. The few works that perform feature learning for \RGBD data
\cite{SocherNIPS2012,bo_iser12} largely ignore the multimodal nature of
the data, not distinguishing the color and depth channels. Here, we present
 a structured regularization approach which allows us to learn more robust
features for \RGBD and other multimodal data.

\subsection{Deep Learning}
Deep learning approaches have demonstrated the ability to \emph{learn} useful
features directly from data for a wide variety of tasks. Early work by
\citet{HintonScience} showed that a deep network trained on images of
hand-written digits will learn features corresponding to pen-strokes. Later
work using localized convolutional features \cite{convNet} showed that these
networks learn features corresponding to object parts when trained on natural
images. This demonstrates that even the basic features learned by these
systems will adapt to the data given. In fact, these approaches are not
restricted to the visual domain, but rather have been shown to learn useful
features for a wide range of domains, such as audio \cite{convDBNAudio,mohamed2012acoustic} and natural language data \cite{NLPfromScratch}.

\header{Deep Learning for Detection:}
However, the vast majority of work in deep learning focuses on classification
problems. Only a handful of previous works have applied these methods 
to detection problems~\cite{osadchy2007synergistic,lecun2004learning,coatesTextDetection}.
For example, \citet{osadchy2007synergistic} and \citet{lecun2004learning} applied a deep energy-based model to the problem of face detection, \citet{sermanet-cvpr-2013} applied a convolutional neural network for pedestrian detection, and
\citet{coatesTextDetection} used a deep learning approach to detect text in
images. \citet{girshick2014rcnn} used learned convolutional features over
image regions for object detection, while \citet{NIPS2013_5207} used a
multi-scale approach based on deep networks for the same task. 

All these approaches focused on object detection and similar problems,
in which the goal is to find a bounding box which tightly contains 
the item to be detected, and for each item, all valid bounding boxes will
be similar. However, in robotic grasp detection, there may be several valid 
grasps for an object in different regions, making it more important to select
the one with the highest chance success. In addition, orientation matters much
more to robotic grasp detection, as most grasps will only be viable for a small
subset of the possible gripper orientations. Our approach to grasp detection
will also generalize across object classes, and even to classes never seen
before by the system, as opposed to the class-specific nature of object
detection.

\begin{figure*}[tbh]
\begin{center}
\begin{minipage}[c]{0.15\textwidth}
\centering
\includegraphics[width=\textwidth]{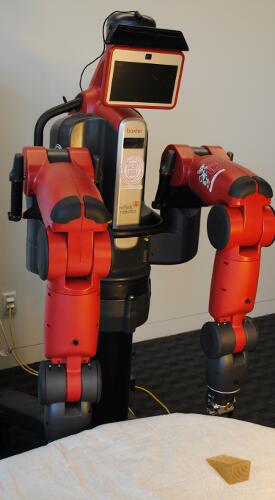}
\end{minipage}
\begin{minipage}[c]{0.15\textwidth}
    \centering
    \includegraphics[width=\textwidth]{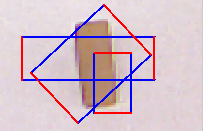} \\ \vspace{0.05in}
\includegraphics[width=\textwidth]{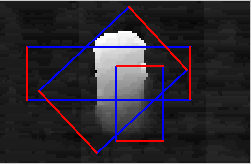}
  \end{minipage}
\begin{minipage}[c]{0.3\textwidth}
\centering
\includegraphics[width=0.3\textwidth]{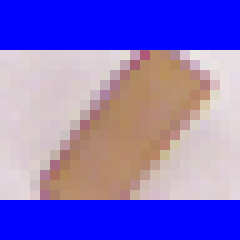}
\includegraphics[width=0.3\textwidth]{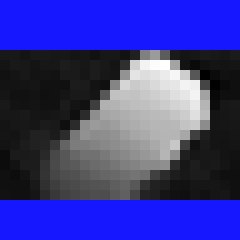}
\includegraphics[width=0.3\textwidth]{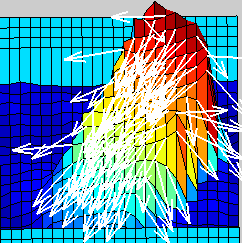}\\ \vspace*{0.05in}
\includegraphics[width=0.3\textwidth]{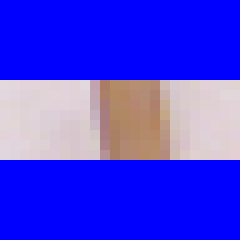}
\includegraphics[width=0.3\textwidth]{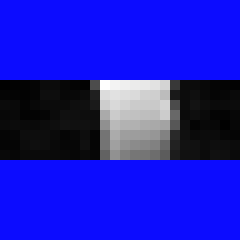}
\includegraphics[width=0.3\textwidth]{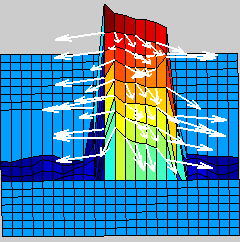}\\ \vspace*{0.05in}
\includegraphics[width=0.3\textwidth]{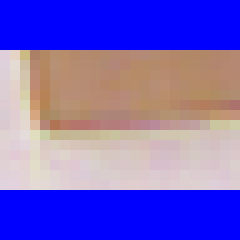}
\includegraphics[width=0.3\textwidth]{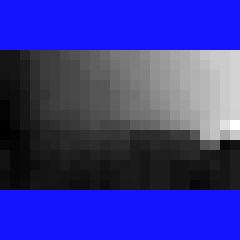}
\includegraphics[width=0.3\textwidth]{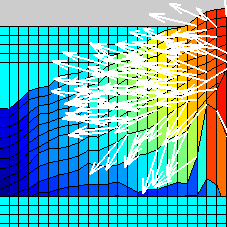}
\end{minipage}
\begin{minipage}[c]{0.07\textwidth}
\includegraphics[width=\textwidth]{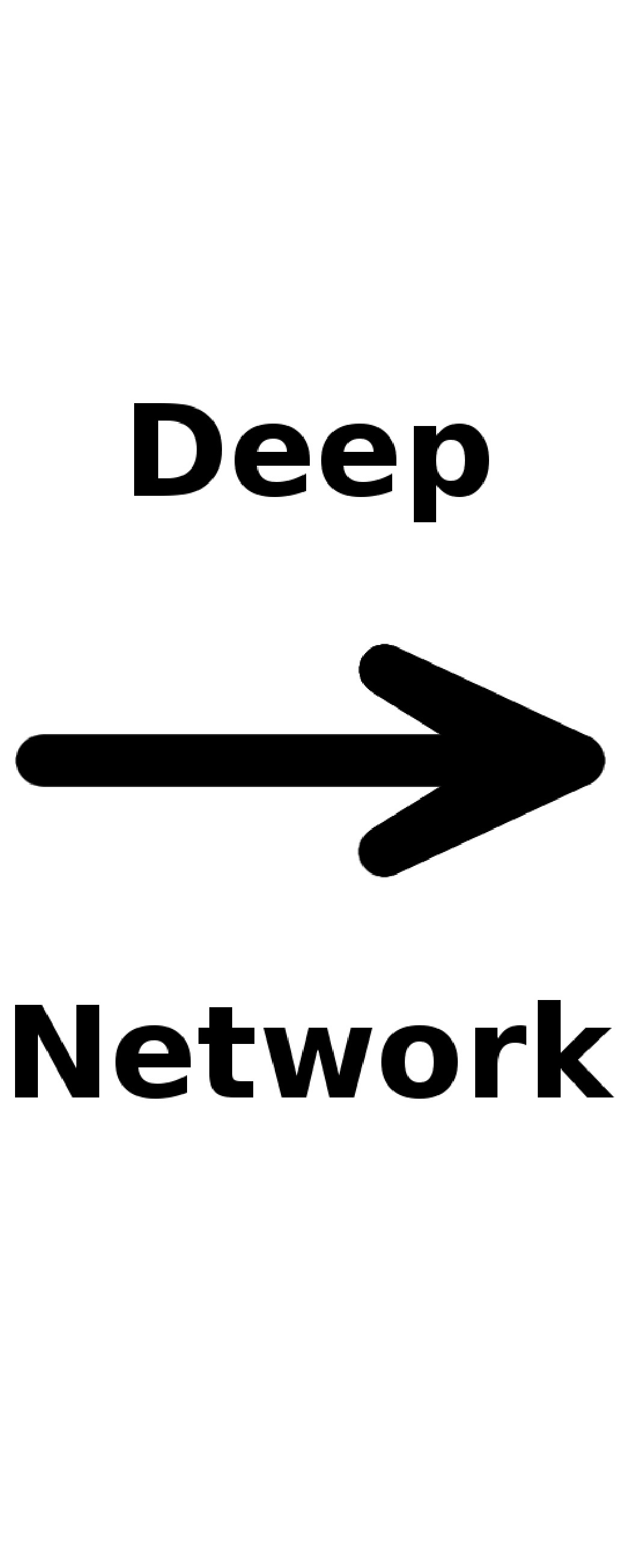}
\end{minipage}
\begin{minipage}[c]{0.25\textwidth}
    \centering
    \includegraphics[width=\textwidth]{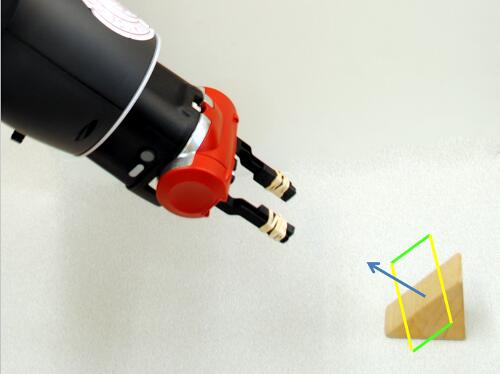}
  \end{minipage}
\end{center}
\vspace*{\captionReduceTop}
\caption{\newTxt{\textbf{Detecting and executing grasps:}
From left to right: Our system obtains an \RGBD image from a Kinect mounted on
the robot, and searches over a large space of possible grasps, for which some
candidates are shown. For each of these, it extracts a set of raw features
corresponding to the color and depth images and surface normals, then uses
these as inputs to a deep network which scores each rectangle. Finally, the
top-ranked rectangle is selected and the corresponding grasp is executed using
the parameters of the detected rectangle and the surface normal at its center.
Red and green lines correspond to gripper plates, blue in \RGBD features
indicates masked-out pixels.}}
\label{fig:graspOverview}
\vspace*{\captionReduceBot}
\vspace*{\figureReduceBot}
\end{figure*}

\header{Multimodal Deep Learning:}
Recent works in deep learning have extended these methods to handle
multiple modalities of input data, such as audio and video \cite{multimodal},
text and image data \cite{srivastava2012multimodal}, and even \RGBD data
\cite{SocherNIPS2012,bo_iser12}.
However, all of these approaches have fallen into
two camps - either learning completely separate low-level features for
each modality \cite{multimodal, srivastava2012multimodal}, or simply concatenating the modalities \cite{SocherNIPS2012,bo_iser12}.
The former approaches have proven effective for 
data where the basic modalities differ significantly, such as the aforementioned
case of text and images, while the latter is more effective in cases where the
modalities are more similar, such as \RGBD data. 

For some new combinations of modalities and tasks, it may not be clear which of
these approaches will give better performance. In fact, in the ideal feature
set, different features may use different subsets of the modalities. 
In this work, we will give a 
structured regularization method which guides the learning algorithm to
select such subsets, without imposing hard constraints on network structure.

\header{Structured Learning and Structured Regularization:}
Several approaches have been proposed which attempt to use a specially-designed
regularization function to impose structure on a set of learned parameters
without directly enforcing it.
\citet{dirtyMulti} used a group regularization function in the multitask
learning setting, where one set of features is used for multiple tasks. This
function applies high-order regularization separately to particular
groups of parameters. Their
function regularized the number of features used for each task in a set of
multi-class classification tasks solved by softmax regression. Intuitively,
this encodes the belief that only some subset of the input features will be
useful for each task, but this set of useful features might vary between tasks.

A few works have also explored the use of structured regularization in deep
learning.
The Topographic ICA algorithm \cite{TICA} is a feature-learning approach that applies a similar penalty term to feature activations, but not to the weights themselves.
\citet{selectRecep} investigate the problem of selecting
receptive fields, i.e., subsets of the input features to be used together in
a higher-level feature. The structure of the network is learned first, then
fixed before learning the parameters of the network.
} 

\begin{figure*}[th!]
\vskip -.05in
\begin{center}
\includegraphics[width=0.9\textwidth]{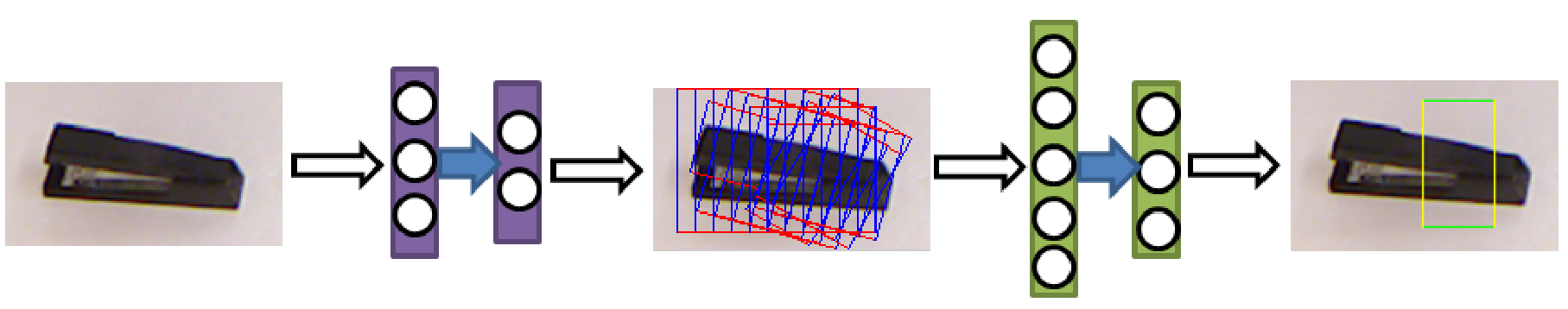}
\end{center}
\vspace*{\captionReduceTop}
\vspace*{\captionReduceTop}
\vspace*{\captionReduceTop}
\caption{\textbf{Illustration of our two-stage detection process:} 
Given an image of
an object to grasp, a small deep network is used to exhaustively search
potential rectangles, producing a small set of top-ranked rectangles. A
larger deep network is then used to find the top-ranked rectangle from these
candidates, producing a single optimal grasp for the given object.}
\vspace*{\captionReduceBot}
\vspace*{\figureReduceBot}
\label{fig:process}
\end{figure*}

\vspace*{\sectionReduceTop}
\section{Deep Learning for Grasp Detection: \\ System and Model\label{sec:detection}}
\vspace*{\sectionReduceBot}
\newTxt{
In this work, we will present an algorithm for robotic grasp detection from
a single \RGBD view. Our approach will be based on machine learning, but
distinguish itself from previous approaches by learning not
only the weights used to rank prospective grasps, but also the
\emph{features} used to rank them, which were previously hand-engineered.

We will do this using deep learning methods, 
learning a set of \RGBD features which will be extracted from each 
candidate grasp, then used to score that grasp. Our approach will include
a structured multimodal regularization method which improves the quality
of the features learned from \RGBD data without constraining network structure.
}

In our system for robotic grasping, as shown in Fig.~\ref{fig:graspOverview},
the robot first obtains an \RGBD image of
the scene containing objects to be grasped. A small deep network is used to score potential
grasps in this image, and a small candidate set of the top-ranked grasps
is provided to a larger deep network, which yields a single best-ranked grasp.

\newTxt{
In this work, we will represent potential grasps using oriented
rectangles in the image plane as seen on the left in 
Fig.~\ref{fig:graspOverview},
with one pair of parallel edges corresponding to
the robotic gripper \cite{JiangICRA2011}. Each rectangle is thus parameterized
by the X and Y coordinates of its upper-left corner, its width, height, and 
orientation in the image plane, giving a five-dimensional search space for
potential grasps. Grasps will be ranked based on features extracted from
the \RGBD image region contained inside their corresponding rectangle, aligned
to the gripper plates, as seen in the center of Fig.~\ref{fig:graspOverview}.
}

To translate a rectangle such as that shown on the right in 
Fig.~\ref{fig:graspOverview}
into a gripper pose for grasping
we find the point with the minimum depth inside the central third (horizontally)
of the rectangle.
We then use the averaged surface normal around this
point to determine the approach vector for the gripper. The orientation of
the detected rectangle is translated to a rotation around this vector to
orient the gripper.
We use the X-Y coordinates of the rectangle center
along with the depth of the closest point to determine a grasping point in 
the robot's coordinate frame. 
We compute a pre-grasp position by shifting 10 cm back from the grasping
point along this approach
vector and position the gripper at this point. We then approach the object 
along the approach vector and grasp it.

Using a standard feature learning approach such as sparse auto-encoder
\cite{SAE}, a deep network can be trained for
the problem of grasping rectangle recognition (i.e., does a given rectangle
in image space correspond to a valid robotic grasp?). However, in a real-world
robotic setting, our system needs to perform \emph{detection} (i.e., given
an image containing an object, how should the robot grasp it?).
This task is significantly more challenging than simple recognition.

\begin{figure}[tb]
\begin{center}
\includegraphics[height=0.2\textheight]{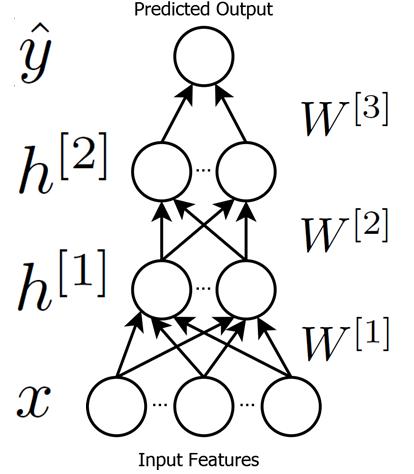}
\hspace{0.25in}
\includegraphics[height=0.15\textheight]{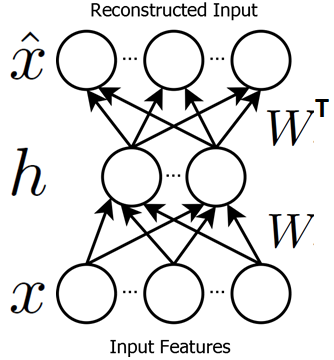}
\end{center}
\caption{\textbf{Deep network and auto-encoder:} 
Left: A deep network with two hidden layers, which
transform the input representation, and a logistic classifier at the top layer,
which uses the features from the second hidden layer to predict the probability
of a grasp being feasible. 
Right: An auto-encoder, used for pretraining. A set of weights
projects input features to a hidden layer. The same weights are then used to
project these hidden unit outputs to a reconstruction of the inputs. In the
sparse auto-encoder (SAE) algorithm, the hidden unit activations are also
penalized.
}
\label{fig:deepNet}
\vspace*{\captionReduceBot}
\vspace*{\figureReduceBot}
\end{figure}

\header{Two-stage Cascaded Detection:}
In order to perform detection,
one naive approach could be to consider each possible oriented rectangle
in the image (perhaps discretized to some level), and evaluate each rectangle
with a deep network trained for recognition.
However, such near-exhaustive search of possible rectangles (based on 
positions, sizes, and orientations) can be quite expensive in practice for
real-time robotic grasping.

Motivated by multi-step cascaded approaches in previous work
\cite{JiangICRA2011,ViolaJonesCascaded},
we instead take a two-stage approach to detection: First, we use a reduced
feature 
set to determine a set of top candidates. Then, 
we use a larger, more robust feature set to rank these candidates. 

However,
these approaches require the design of two separate sets of features. In
particular, it can be difficult to manually design a small set of first-stage features 
which is both quick to compute and robust enough to produce a good set of
candidate detections for the second stage. Using deep learning 
allows us to 
circumvent the costly manual design of features by simply training networks 
of two different sizes, using the smaller for the 
exhaustive first pass, and the
larger to re-rank the candidate detection results.



\newcommand{\maskchar}{\mu}
\newcommand{\scalechar}{\Psi}
\newcommand{\scalecharcase}{\psi}

\header{Model:}
To detect robotic grasps from the rectangle representation, we model the
probability of a rectangle $G^{(t)}$, with features $x^{(t)} \in \mathbb{R}^N$
 being graspable, using a
random variable $\hat{y}^{(t)}\in\{0,1\}$ which indicates whether or not we
predict $G^{(t)}$ to be graspable. We use a deep network, as shown in
Fig.~\ref{fig:deepNet}-left, with 
two layers of sigmoidal
hidden units $h^{[1]}$ and $h^{[2]}$, with $K_1$ and $K_2$ units per layer,
respectively. A logistic classifier over the outputs of the second-layer 
hidden units
then predicts $P(\hat{y}^{(t)} | x^{(t)}; \Theta)$, \newTxt{so chosen because
ground-truth graspability is represented as binary}. 
Each layer $\ell$ will have a set of weights $W^{[\ell]}$
mapping from its inputs to its hidden units, so the parameters of our model are
$\Theta = \{W^{[1]}, W^{[2]}, W^{[3]} \}$. Each hidden unit forms output
by a sigmoid $\sigma(a) = 1/(1+\exp(-a))$ over its weighted input:
\begin{align}
h_j^{[1](t)} &= \sigma \left( \sum_{i=1}^N x_i^{(t)} W^{[1]}_{i,j}  \right)   \displaybreak[0] \nonumber\\
h_j^{[2](t)} &= \sigma \left( \sum_{i=1}^{K_1} h_i^{[1](t)} W^{[2]}_{i,j}  \right) \displaybreak[0] \nonumber\\
P(\hat{y}^{(t)} = 1 | x^{(t)}; \Theta)& = \sigma \left( \sum_{i=1}^{K_2} h_i^{[2](t)} W^{[3]}_{i}  \right)
\end{align}

\vspace*{\subsectionReduceTop}
\subsection{Inference and Learning}
\vspace*{\subsectionReduceBot}
During \textbf{inference}, our goal is to find the single grasping rectangle
with the maximum probability of being graspable for some new object.
With $G$ representing a particular grasping rectangle position, orientation,
 and size, we find this best rectangle as:
\begin{align}
G^* &= \underset{G}{\mbox{arg max }}P(\hat{y}^{(t)} = 1 | \phi(G); \Theta)
\end{align}
Here, the function $\phi$ extracts the appropriate input representation
for rectangle $G$.

During \textbf{learning}, our goal is to learn the parameters
$\Theta$ that optimize the recognition accuracy of our system.
Here, input data is given as a set of pairs of features 
$x^{(t)} \in \mathbb{R}^N$ and
ground-truth labels $y^{(t)} \in \{0,1\}$ for $t = 1,\ldots,M$. 
 As in most
deep learning works, we  use a two-phase learning approach.


In the first phase, we will use \emph{unsupervised feature learning} to initialize the
hidden-layer weights $W^{[1]}$ and $W^{[2]}$. Pre-training weights this way
is critical to avoid overfitting. We will use a variant of a sparse
auto-encoder (SAE) \cite{SAE}, as illustrated in 
Fig.~\ref{fig:deepNet}-right. 
We define $g(h)$ as a sparsity penalty  function over hidden unit  activations, with $\lambda$ controlling its weight. 
With $f(W)$ as a regularization function, weighted by $\beta$, and
$\hat{x}^{(t)}$ as the reconstruction of $x^{(t)}$, SAE solves the following
to initialize hidden-layer weights:
\begin{align}
W^* & =  \underset{W}{\mbox{arg min }} \sum_{t=1}^M ( ||\hat{x}^{(t)} - x^{(t)}||_2^2
+ \lambda \sum_{j=1}^K g(h_j^{(t)}) ) + \beta f(W) \label{eqn:sparseAuto} \nonumber\\
h_j^{(t)} &= \sigma ( \sum_{i=1}^N x_i^{(t)} W_{i,j} )  \nonumber\\
\hat{x}_i^{(t)} &= \sum_{j=1}^K h_j^{(t)} W_{i,j} 
\end{align}
We first use this algorithm to initialize $W^{[1]}$ to reconstruct $x$. 
We then fix $W^{[1]}$ and
learn $W^{[2]}$ to reconstruct $h^{[1]}$.

During the \emph{supervised} phase of the learning algorithm,
we then jointly learn classifier
weights $W^{[3]}$ and fine-tune hidden layer weights $W^{[1]}$ and $W^{[2]}$
for recognition.
We maximize the log-likelihood of the data along with regularization penalties
on hidden layer weights:
\begin{align}
\Theta^* & = \underset{\Theta}{\mbox{arg max }} \sum_{t=1}^M 
\log P(\hat{y}^{(t)} = y^{(t)} | x^{(t)}; \Theta) \nonumber \\
& \qquad\qquad\qquad - \beta_1 f(W^{[1]})
- \beta_2 f(W^{[2]}) \label{eqn:backprop}
\end{align}

\header{Two-stage Detection Model:}
 During \textbf{inference} for two-stage detection, 
we will first use a smaller network
to produce a set of the top $T$ rectangles with the highest probability of being
graspable according to network parameters $\Theta_1$. We will then use a larger
network with a separate set of parameters $\Theta_2$ to re-rank these $T$
rectangles and obtain a single best one. The only change to \textbf{learning}
for the two-stage model is that these two sets of parameters are learned
separately, using the same approach.

\section{System Details}\label{sec:input}
\newTxt{
In this section, we will define the set
of raw features which our system will use, forming $x$ in the equations above,
and how they are extracted from an \RGBD image. Some examples of these
features are shown in Fig~\ref{fig:graspOverview}.

Our algorithm uses only local information - specifically, we extract the \RGBD
sub-image contained within each rectangle, and use this to generate features
for that rectangle. This image is rotated so that its left and right edges
correspond to the gripper plates, and then re-scaled to fit inside the network's
receptive field.

From this 24x24 pixel image, seven channels' worth of features are extracted,
giving 24x24x7 = 4032 input features. The first three channels are the image in
YUV color space, used because it represents
image intensity and color separately. The next is simply the depth channel of
the image. The last three are the X, Y, and Z components of surface normals
computed based on the depth channel. These are computed after the
image is aligned to the gripper so that they are always relative to the gripper plates.}

\begin{figure}[tb]
\begin{center}
\includegraphics[width=.3\linewidth]{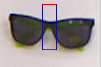} \hspace{0.3in}
\includegraphics[height=0.7in]{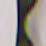} \hspace{0.3in}
\includegraphics[height=0.7in]{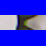}
\end{center}
\vspace*{-0.07in}
\vspace*{\captionReduceTop}
\caption{\textbf{Preserving aspect ratio:} Left: a pair of sunglasses
with a potential grasping rectangle. Red edges indicate gripper plates.
Center: image taken from the rectangle and rescaled to fit a square
aspect ratio. Right: same image, padded and centered in the
receptive field. Blue areas indicate masked-out padding. When rescaled,
the rectangle incorrectly appears graspable. Preserving aspect ratio and
padding allows the rectangle to correctly appear non-graspable.}
\vspace*{\captionReduceBot}
\vspace*{\figureReduceBot}
\label{fig:aspectRatio}
\end{figure}

\begin{figure}[t]
\begin{center}
\includegraphics[height=0.1\textheight]{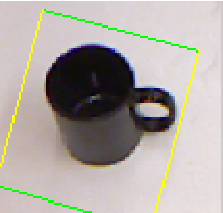}
\includegraphics[height=0.1\textheight]{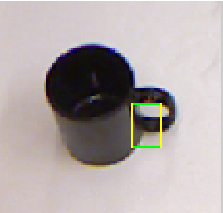}
\end{center}
\vspace*{\captionReduceTop}
\caption{\textbf{Improvement from mask-based scaling:} Left: Result without mask-based scaling. Right: Result with mask-based scaling.}
\vspace*{\captionReduceBot}
\vspace*{\figureReduceBot}
\label{fig:enNorm}
\end{figure}

\subsection{Data Pre-Processing}
Whitening data is critical for deep learning approaches to work well,
especially in cases such as multimodal data where the statistics of the
input data may vary greatly. While PCA-based approaches have been shown
to be effective \cite{pcaWhiten}, they are difficult to apply in cases such
as ours where large portions of the data may be masked out.

Depth data, in particular, can be difficult to whiten because the range
of values may be very different for different patches in the image. Thus,
we first whiten each depth patch individually, subtracting the patch-wise
mean and dividing by the patch-wise standard deviation, down to some 
minimum.

For multimodal data,
the statistics of the data for each modality should match as closely as
possible, to avoid learning features which are biased towards or away from using
particular modes. This is particularly important when regularizing 
each modality
separately, as in our approach. Thus, we drop mean values for each feature
separately, but scale the data for each channel by dividing by the standard 
deviation of all its features combined. 

\begin{figure*}[tb]
\begin{center}
\includegraphics[width=.9\textwidth]{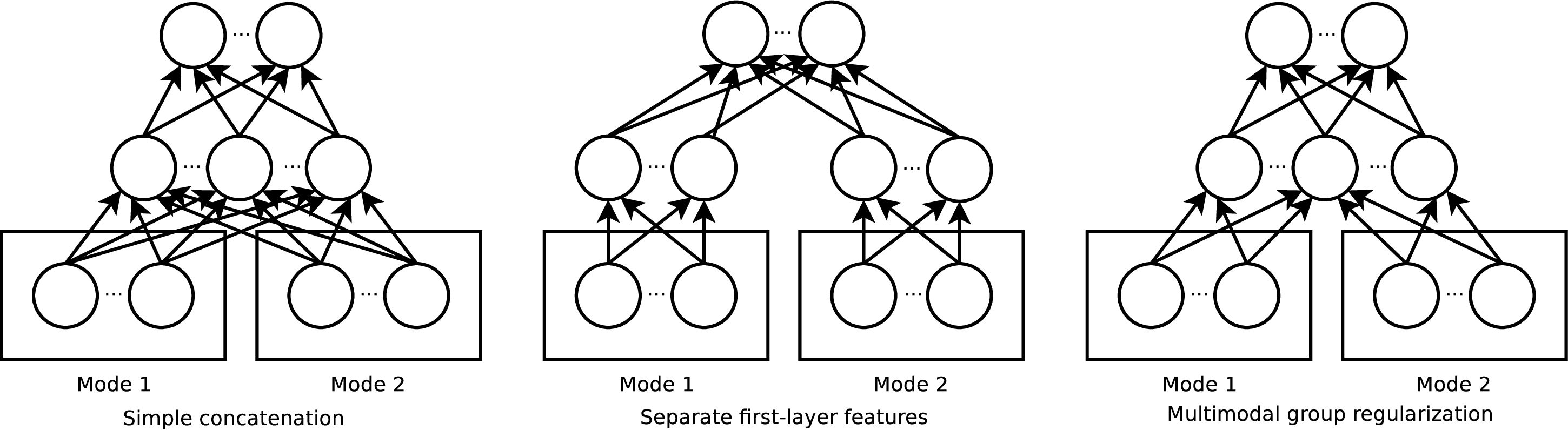}
\end{center}
\vspace{\captionReduceTop}
\caption{\textbf{Three possible models for multimodal deep learning:} 
Left: fully dense model---all visible features are concatenated and
modality information is ignored. Middle: modality-specific sparse model - separate first layer
features are trained for each modality. Right: group-sparse model---a structured regularization term encourages features to use only a
subset of the input modes.}
\vspace*{\captionReduceBot}
\vspace*{\figureReduceBot}
\label{fig:multiModels}
\end{figure*}

\subsection{Preserving Aspect Ratio.}

It is important for to preserve aspect ratio when feeding features 
into the network. This is because distorting image features may cause
non-graspable rectangles to appear graspable, as shown in 
Fig.~\ref{fig:aspectRatio}.
\newTxt{
However, padding with zeros can cause rectangles with less padding to receive
higher graspability scores, as the network will have more nonzero inputs. 
It is important to account
for this because in many cases the ideal grasp for an object might be
represented by a thin rectangle which would thus contain many zero values
in its receptive field from padding.}

To address this problem,
we scale up the magnitude of the available input for each rectangle based on
the fraction of the rectangle which is masked out. In particular, we define 
a multiplicative scaling factor for the inputs
from each modality, based on the fraction of each mode
which is masked out, since each mode may have a different mask.

In the multimodal setting, we assume that the input data $x$ is known to come from $R$ distinct
modalities, for example audio and video data, or depth and RGB data. We
define the modality matrix $S$ as an $R$x$N$ binary 
matrix, where each element $S_{r,i}$ indicates membership of visible unit
$x_i$ in a particular modality $r$, such as depth or image intensity. The
scaling factor for mode $r$ is then defined as:
$\scalechar_r^{(t)} =  \sum_{i=1}^N S_{r,i}/\left(\sum_{i=1}^N S_{r,i} \maskchar_i^{(t)}\right)$,
where $\maskchar_i^{(t)}$ is 1 if $x_i^{(t)}$ is masked in, 
0 otherwise. The scaling factor for case $i$ is:
$\scalecharcase_i^{(t)} = \sum_{r=1}^R S_{r,i} \scalechar_r^{(t)}$.



We could simply scale up each value of $x$ by its corresponding scale factor
when training our model, as ${x'}_i^{(t)} = \scalecharcase_{i}^{(t)} x_i^{(t)}$.
However, since our sparse autoencoder penalizes squared error, scaling $x$
linearly will scale the error for the corresponding cases quadratically,
causing the learning algorithm to lend increased significance to cases
where more data is masked out. Instead, we can use the scaled $x'$ as input
to the network, but penalize reconstruction based on the original $x$, only
scaling after the squared error has been computed:
\begin{align}
W^* & = \underset{W}{\mbox{arg min}} \sum_{t=1}^M \left( \sum_{i=1}^N 
\scalecharcase_i^{(t)} (\hat{x}_i^{(t)} - x_i^{(t)})^2 
+ \lambda \sum_{j=1}^K g(h_j^{(t)}) \right) \label{eqn:sparseAutoEnNorm}
\end{align}
We redefine the hidden units to use the scaled visible input:
\begin{align}
h_j^{(t)} &= \sigma \left(\sum_{i=1}^N {x'}_i^{(t)} W_{i,j} \right)
\end{align}
This approach is equivalent to adding additional, potentially fractional, 
`virtual' visible units to the model based on the scaling factor for each
mode. In practice, we found it necessary to limit the scaling factor to 
a maximum of some value $c$, as ${\scalechar'}_r^{(t)} = \mbox{min}(\scalechar_r^{(t)},c)$.

As shown in Table~\ref{tbl:graspDet} our mask-based scaling 
technique at the visible layer improves
grasping results by over 25\% for both metrics.  As seen in 
Figure~\ref{fig:enNorm}, it removes the network's inherent
bias towards square rectangles, exhibiting a much wider range of aspect
ratios that more closely matches that of the ground-truth data.

\vspace*{\subsectionReduceTop}
\section{Structured Regularization for \\ Feature Learning\label{sec:sparsity}}
\vspace*{\subsectionReduceBot}

\begin{figure*}[tbh!]
\begin{center}
\subfigure[Features corresponding to positive grasps.]{
\includegraphics[width=0.45\textwidth]{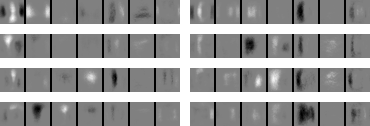}
\vspace{-0.05in}
}
\hspace*{0.25in}
\subfigure[Features corresponding to negative grasps.]{
\includegraphics[width=0.45\textwidth]{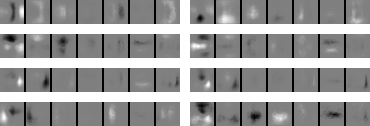}
\vspace{-0.05in}
}
\end{center}
\vspace{\captionReduceTop}
\vspace{\captionReduceTop}
\caption{\textbf{Features learned from grasping data:} Each feature contains seven channels - from
left to right, depth, Y, U, and V image channels, and X, Y, and Z 
surface normal components.
 Vertical edges correspond to gripper plates.
Left: eight features with the strong positive correlations to rectangle graspability.
Right: similar, but negative correlations. Group regularization
eliminates many modalities from many of these features, making them more robust.
}
\vspace*{\captionReduceBot}
\vspace*{\figureReduceBot}
\label{fig:graspFeat}
\end{figure*}

 A naive way
of applying feature learning to multimodal data is to simply take $x$ (as a concatenated vector) as input to the model described above, ignoring information about specific modalities, as
seen on the lefthand side of Figure~\ref{fig:multiModels}. This approach
may either 1) prematurely learn features which include all modalities, which can lead to
overfitting,
or 2) fail to learn associations between modalities with very different underlying statistics. 

Instead of concatenating multimodal input as a vector, Ngiam et al. \cite{multimodal} proposed training a first layer representation for
each modality separately, as shown in Figure~\ref{fig:multiModels}-middle. This
approach makes the assumption that the ideal low-level features for each
modality are purely unimodal, while higher-layer features are purely multimodal. 
This approach may work better for some problems where
the modalities have very different basic representations, such as the video
and audio data (as used in \cite{multimodal}), so that separate first layer features may give
better performance. However, for modalities such as \RGBD data, where the input
modes represent different channels of an image, learning low-level correlations
can lead to more robust features -- our experiments in 
Section~\ref{sec:experiments} show that simply 
concatenating the input modalities significantly outperforms training
separate first-layer features for robotic grasp detection from \RGBD data.

For many problems, it may be difficult to tell which of these approaches will
perform better, and time-consuming to tune and comparatively evaluate multiple
algorithms. In addition, the ideal feature set for some problems may contain
features which use some, but not all, of the input modalities, a case which
neither of these approaches are designed to handle.

To solve these problems, we propose a new algorithm for feature learning for
multimodal data. 
Our approach incorporates a 
structured penalty term into the optimization problem to be solved during learning. This technique
allows the model to learn correlated features between multiple input modalities,
but regularizes the number of modalities used per feature (hidden unit), 
discouraging the
model from learning weak correlations between modalities. 
With this regularization term, the algorithm can specify how mode-sparse or mode-dense the features should be, representing a continuum between the two extremes outlined above.

\header{Regularization in Deep Learning:}
In a typical deep learning model, \Ltwo {} regularization (i.e., $f(W) = ||W||_2^2$) or \Lone {} regularization (i.e., $f(W) = ||W||_1$) are commonly used in training (e.g., as specified in Equations \eqref{eqn:sparseAuto} and \eqref{eqn:backprop}).
 These are often called a ``weight cost" 
(or ``weight decay"), and are left implicit in many works.

Applying regularization is well known to improve the generalization
performance of feature learning algorithms. One might expect that a simple
\Lone {} penalty would eliminate weak correlations in multimodal features, leading
to features which use only a subset of the modes each. However, we found 
that in
practice, a value of $\beta$ large enough to cause this also degraded the
quality of features for the remaining modes and lead to decreased task
performance. 

\header{Multimodal Regularization:}
Structured regularization, such as in \cite{dirtyMulti},
takes a set of groups of
weights, and applies some regularization function (typically high-order) 
\emph{separately} to each group.
In our structured multimodal regularization algorithm, each modality 
will be used as a regularization group separately
for each hidden unit. For example, a group-wise p-norm would
be applied as:
\begin{align}
f(W) & = \sum_{j=1}^K \sum_{r=1}^R \left( \sum_{i=1}^N S_{r,i} | W_{i,j}^p |
\right)^{1/p}
\end{align}
where $S_{r,i}$ is 1 if feature $i$ belongs to group $r$ and 0 otherwise.
Using a high value of $p$ allows us to penalize higher-valued weights from each
mode to each feature more strongly than lower-valued ones. This also means that
forming a high-valued weight in a group with other high-valued weights will
accrue a lower additional penalty than doing so for a group with only low-valued
weights. At the limit ($p\rightarrow \infty$), this group regularization becomes equivalent to the infinity (or max) norm:
%
\begin{align}
f(W) & = \sum_{j=1}^K \sum_{r=1}^R \max_{i} S_{r,i} |W_{i,j}|
\end{align}
which penalizes only the maximum weight from each mode to each feature. In
practice, the infinity norm is not differentiable and therefore is difficult to apply gradient-based optimization methods; 
in this paper, we use the log-sum-exponential as a differentiable approximation to the max norm.

In experiments, this regularization function produces first-layer weights
concentrated in fewer modes per feature. 
However, we found that at values of $\beta$ sufficient to induce the desired
mode-wise sparsity patterns, penalizing the maximum also had the undesirable
side-effect of causing many of the weights for other modes to saturate at 
their mode's maximum, suggesting that the features were overly constrained.
In some cases, constraining the weights in this manner also caused
the algorithm to learn duplicate (or redundant) features, in effect scaling up the
feature's contribution to reconstruction to compensate for its constrained
maximum. This is obviously an undesirable effect, as it reduces the effective
size (or diversity) of the learned feature set. 


This suggests that the max-norm may be overly constraining. A more desirable 
sparsity function would penalize nonzero weight maxima for each mode for each
feature without additional penalty for larger values of these maxima. We can
achieve this effect by applying the \Lzero {} norm, which takes a value of 0 for
an input of 0, and 1 otherwise, on top of the max-norm from above:
\begin{align}
f(W) & = \sum_{j=1}^K \sum_{r=1}^R \mathbb{I}\{(\max_{i} S_{r,i} |W_{i,j}|) > 0\}
\end{align}
where $\mathbb{I}$ is the indicator function, which takes a value of 1 if its
argument is true, 0 otherwise. 
Again, for a gradient-based method, we used an approximation to the
 \Lzero {} norm, such as $\log(1+x^2)$. This regularization function now encodes a
 direct penalty on the number of modes used for each weight, without further
constraining the weights of modes with nonzero maxima.

Figure~\ref{fig:graspFeat} shows features learned from the unsupervised stage
of our group-regularized deep learning algorithm. We discuss these features,
and their implications for robotic grasping, in Section~\ref{sec:results}.


\vspace{\sectionReduceTop}
\section{Experiments\label{sec:experiments}}
\vspace{\sectionReduceBot}

\begin{figure}
\begin{center}
\includegraphics[width=\linewidth]{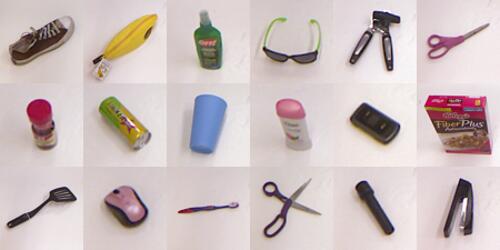}
\end{center}
\vspace*{\captionReduceTop}
\caption{\textbf{Example objects from the Cornell grasping dataset:} \cite{JiangICRA2011}.  This dataset contains 
objects from a large variety of categories.}
\vspace*{\captionReduceBot}
\vspace*{\figureReduceBot}
\label{fig:datasetObj}
\end{figure}

\subsection{Dataset}
We used the extended version of the Cornell grasping
dataset for our experiments. This dataset, along with code for this
paper, is available at 
\texttt{\url{http://pr.cs.cornell.edu/deepgrasping}}.
We note that this is an updated version of the dataset used in
 \cite{JiangICRA2011}, containing several more complex objects, and thus
results for their algorithms will be different from those in  
\cite{JiangICRA2011}.
This dataset contains 1035 images of 280 graspable objects, several of
which are shown in Fig.~\ref{fig:datasetObj}. 
Each image is annotated with several ground-truth positive and negative
grasping rectangles. While the vast majority of possible rectangles for most objects
will be non-graspable, the dataset contains roughly equal numbers of
graspable and non-graspable rectangles. We will show that this is useful
for an unsupervised learning algorithm, as it allows learning a good 
representation for graspable rectangles even from unlabeled data. 

We performed five-fold cross-validation, and present results for splits on per image 
(i.e., the training set and the validation set do not share the same image) and per
object (i.e., the training set and the validation set do not share any images from the same object) basis. \newTxt{Hyper-parameters were selected by validating
performance on a separate set of 300 grasps not used in any of the 
cross-validation splits.}

\newTxt{We take seven 24x24 pixel channels as described in 
Section~\ref{sec:input} as
input, giving 4032 input features to each network.}
We trained a deep network with 200 hidden units each at the first and second layers
using our learning algorithm as described in Sections~\ref{sec:detection} 
and~\ref{sec:sparsity}. Training this network took roughly 30 minutes.
For trials involving our two-pass system, we trained
a second network with 50 hidden units at each layer in the same manner.
During inference we performed an exhaustive search using this network, then
used the 200-unit network to re-rank the 100 highest-ranked rectangles found
by the 50-unit network.


\subsection{Baselines}
We compare our recognition results in the Cornell grasping dataset with
the features from \cite{JiangICRA2011}, as well as the combination of these
features and Fast Point Feature Histogram (FPFH) features \cite{FPFH}. 
We used a linear SVM for classification, which gave the best results among all other kernels. We also report chance performance, obtained by
randomly selecting a label in the recognition case, and randomly assigning
scores to rectangles in the detection case.

We also compare our algorithm to other deep learning approaches. We compare
to a network trained only with standard L1 regularization, and a network
trained in a manner similar to \cite{multimodal}, where three separate sets
of first layer features are learned for the depth channel, the combination
of the Y, U, and V channels, and the combination of the X, Y, and Z surface
normal components.



\subsection{Metrics for Detection}
For detection, we compare the top-ranked rectangle for each method with the 
set of ground-truth rectangles for each image. We present results using two metrics, the ``point'' and ``rectangle'' metric.

For the point metric, similar to \citet{Saxena-IJRR-Grasping},
 we compute the center point of the predicted rectangle,
and consider the grasp a success if it is within some distance from at least
one ground-truth rectangle center. We note that this metric ignores grasp
orientation, and therefore might overestimate the performance of an algorithm
for robotic applications.

For the rectangle metric, similar to \citet{JiangICRA2011}, 
let $G$ be the top-ranked
grasping rectangle predicted by the algorithm, and $G^*$ be a ground-truth
rectangle. Any rectangles with an orientation error of more than $30^o$ from
$G$ are rejected. From the remaining set, we use the common bounding box
evaluation metric of intersection divided by union - i.e. 
$Area(G\cap G^*)/Area(G\cup G^*)$. Since a ground-truth rectangle can define
a large space of graspable rectangles (e.g., covering the entire length of a pen),
we consider a prediction to be correct if it scores at least 25\% by this 
metric.



\vspace*{\sectionReduceTop}
\section{Results and Discussion \label{sec:results}}
\vspace*{\sectionReduceBot}

\newcommand{\meshWidth}{0.22\linewidth}

\begin{figure*}[tb]
\begin{center}
\begin{tabular}{ccccc}
\raisebox{.25in}{\rotatebox{90}{\small {Positive}}} & 
\includegraphics[width=\meshWidth, clip=true, trim=0 45 0 50]{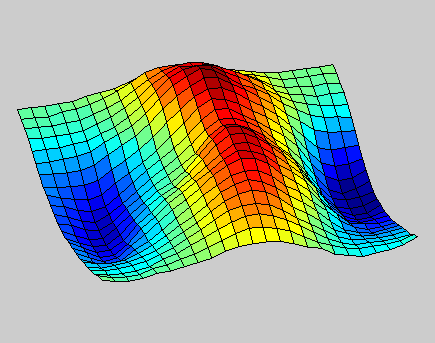} & 
\includegraphics[width=\meshWidth, clip=true, trim=0 45 0 50]{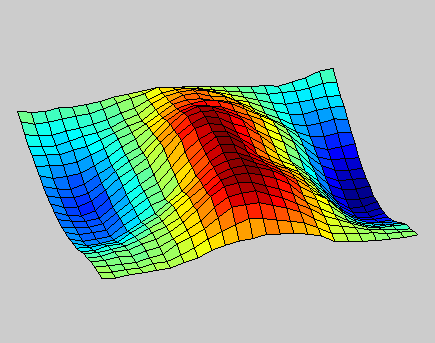} & 
\includegraphics[width=\meshWidth, clip=true, trim=0 35 0 60]{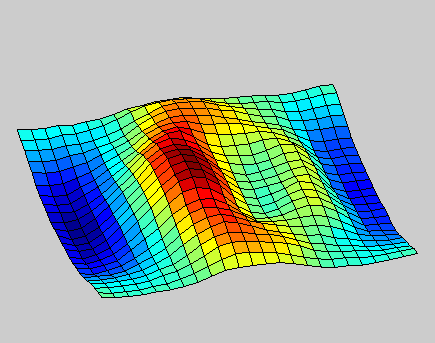} & 
\includegraphics[width=\meshWidth, clip=true, trim=0 45 0 50]{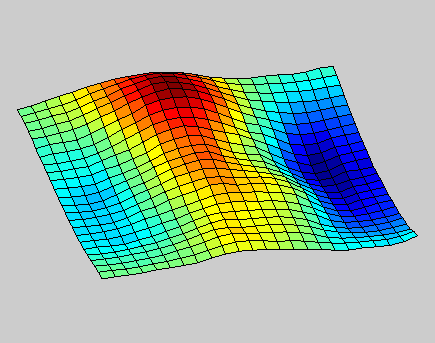} \\
\raisebox{.25in}{\rotatebox{90}{\small{Negative}}} & 
\includegraphics[width=\meshWidth, clip=true, trim=0 35 0 60]{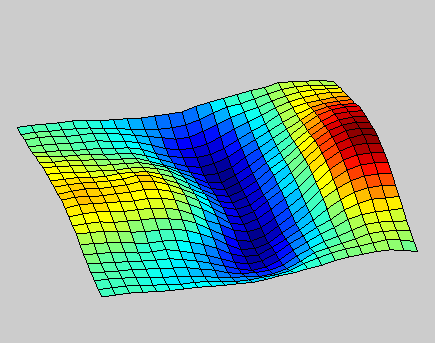} & 
\includegraphics[width=\meshWidth, clip=true, trim=0 60 0 35]{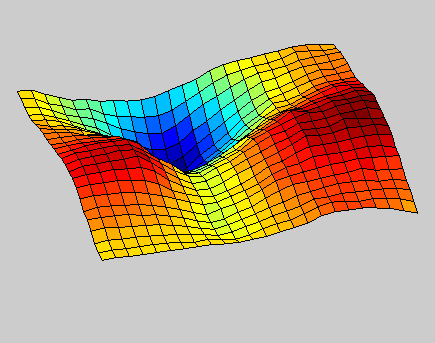} & 
\includegraphics[width=\meshWidth, clip=true, trim=0 60 0 35]{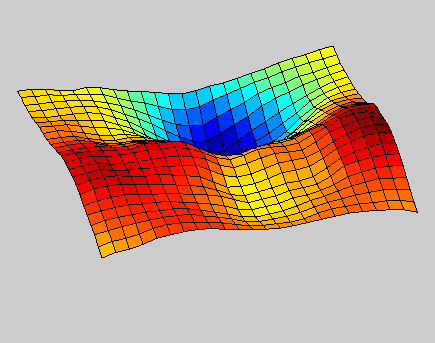} & 
\includegraphics[width=\meshWidth, clip=true, trim=0 65 0 30]{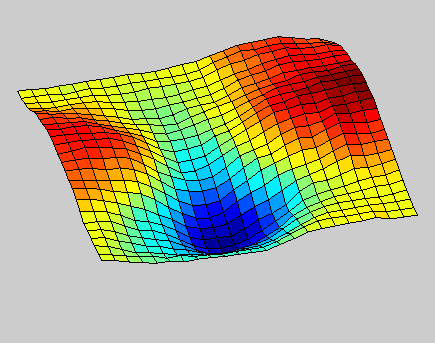}
\end{tabular}
\end{center}
\vspace{-0.1in}
\vspace{\captionReduceTop}
\caption{\textbf{Learned 3D depth features:} 3D meshes for depth channels of the four features with strongest positive (top) and negative(bottom) correlations to rectangle graspability. 
Here X and Y coordinates corresponds to positions in the deep network's receptive field, and Z coordinates corresponds to weight values to the depth channel for each location.
Feature shapes clearly correspond to graspable and
non-graspable structures, respectively.}
\vspace*{\captionReduceBot}
\vspace*{\figureReduceBot}
\label{fig:meshes}
\end{figure*}

\begin{table}[tb]
\begin{center}
\caption{\textbf{Recognition results for Cornell grasping dataset.} }
\label{tbl:graspRec}
{\footnotesize 
\begin{tabular}{l | r}
\whline{\tabTopBotLnWd}
Algorithm & Accuracy (\%)\\ \hline
Chance & 50 \\
\citet{JiangICRA2011} & 84.7\\
\citet{JiangICRA2011} + FPFH & 89.6\\
Sparse AE, separate layer-1 feat. & 92.8\\
Sparse AE & \textbf{93.7}\\
Sparse AE, group reg. & \textbf{93.7}\\
\whline{\tabTopBotLnWd}
\end{tabular}}
\end{center}
\vspace*{\captionReduceBot}
\vspace*{\figureReduceBot}
\end{table}

\subsection{Deep Learning for Robotic Grasp Detection}
\vspace*{\subsectionReduceBot}
Figure~\ref{fig:graspFeat} shows the features learned by the unsupervised phase of
our algorithm which have
a high correlation to positive and negative grasping cases.
Many of these features show non-zero weights to the depth channel, 
indicating that it learns the 
correlation of depths to graspability.
We can see that weights to many of the modalities for these features have been
eliminated by our structured regularization approach. In particular,
many of these features lack weights to the U and V ($3^{rd}$ and $4^{th}$)
channels, which correspond to color, allowing the system to be more robust
to different-colored objects.

Figure~\ref{fig:meshes} shows 3D meshes for the depth channels of the
four features with the strongest
positive and negative correlations to valid grasps. 
Even \emph{without any
supervised information}, our algorithm was able to learn several features which
correlate strongly to graspable cases and non-graspable cases.
The first two positive-correlated features represent
handles, or other cases with a raised region in the center, while the
second two represent circular rims or handles. The negatively-correlated
features represent obviously non-graspable cases, such as ridges perpendicular
to the gripper plane and ``valleys'' between the gripper plates. 
From these features, we can see that even during unsupervised feature learning, 
our approach is able to learn a representation useful for the task at hand,
thanks purely to the fact that the data used is composed of half graspable
and half non-graspable cases.

From Table~\ref{tbl:graspRec}, we see that the recognition performance is 
significantly improved with deep learning methods,
improving 9\% over the features from \cite{JiangICRA2011} and 4.1\% over
those features combined with FPFH features. Both \Lone {} and group regularization
performed similarly for recognition, but training separate first layer
features decreased performance slightly. This shows that learned features,
in addition to avoiding hand-design, are able to improve performance
significantly over the state of the art. It demonstrates that a deep network
is able to learn the concept of ``graspability'' in a way that generalizes
to new objects it hasn't seen before. 

Table~\ref{tbl:modes} shows that even
using any one of the three input modalities (RGB, depth, or surface normals),
our algorithm is able to learn features which outperform hand-engineered ones
for recognition. Depth gives the highest performance of any single-mode network.
Combining depth and normal information improves results over either alone,
indicating that they give non-redundant information. 

\begin{table}[tb]
\begin{center}
\caption{\textbf{Recognition results for different modalities,} for a deep
network pre-trained using SAE.}
\label{tbl:modes}
{\footnotesize 
\begin{tabular}{l | r}
\whline{\tabTopBotLnWd}
Modes & Accuracy (\%)\\ \hline
Chance & 50 \\
RGB & 90.3 \\
Depth & 92.4 \\
Surf. Normals & 90.3 \\
Depth + Surf. Normals & 92.8 \\
RGB + Depth + Surf. Normals & \textbf{93.7}\\
\whline{\tabTopBotLnWd}
\end{tabular}}
\end{center}
\vspace*{\captionReduceBot}
\vspace*{\figureReduceBot}
\end{table}

The highest accuracy
is still obtained by using all the input modalities. This shows that combining 
depth and color information leads to a system which is more robust than either
modality alone. This is due to the fact that some graspable cases (rims of
monochromatic objects, etc.) can only be detected using depth information,
while in others, the depth channel may be extremely noisy, requiring the
use of color information. From this, we can see that integrating multimodal
information, a major focus of this work, is important in recognizing good
robotic grasps. 

Table~\ref{tbl:graspDet} shows that the performance gains from deep learning
for recognition carry over to detection, as well.
Once mask-based scaling has been applied, all deep learning approaches except
for training separate first-layer features outperform the hand-engineered
features from \cite{JiangICRA2011} by up to 13\% for the point metric and
17\% for the rectangle metric, while also avoiding the need to design
task-specific features. Without mask-based scaling, the system performs
poorly, due to the bias illustrated in Fig.~\ref{fig:enNorm}.
Separate first-layer features also give weak detection performance, 
indicating that the relative scores assigned by this form of network
are less robust than those learned using our structured regularization
approach.

\begin{table}[tb]
\begin{center}
\caption{\textbf{Detection results for point and rectangle metrics},
for various learning algorithms.
}
\label{tbl:graspDet}
{\footnotesize
\begin{tabular}{l|c|c|c|c}
\whline{\tabTopBotLnWd}
\multirow{2}{*}{Algorithm} & \multicolumn{2}{c |}{Image-wise split} & \multicolumn{2}{c}{Object-wise split}\\ 
\cline{2-5}
& Point & Rect  & Point   & Rect  \\ 
\hline
Chance & 35.9 & 6.7 & 35.9 & 6.7 \\
\citet{JiangICRA2011} & 75.3 & 60.5 & 74.9 & 58.3 \\
SAE, no mask-based scaling & 62.1 & 39.9 & 56.2 & 35.4 \\
SAE,  separate layer-1 feat. &  70.3 & 43.3 & 70.7 & 40.0 \\
SAE, \Lone~reg.  & 87.2 & 72.9 & \textbf{88.7} & 71.4 \\
SAE,  struct.~reg., 1$^{st}$ pass only & 86.4 & 70.6 & 85.2 & 64.9\\
SAE,  struct.~reg., 2$^{nd}$ pass only & 87.5 & 73.8 & 87.6 & 73.2\\
SAE, struct.~reg.~two-stage & \textbf{88.4} & \textbf{73.9} & 88.1 & \textbf{75.6} \\
\whline{\tabTopBotLnWd}
\end{tabular}}
\end{center}
\vspace*{\captionReduceBot}
\vspace*{\figureReduceBot}
\end{table}
Using structured multimodal regularization also improves results over
standard \Lone regularization by up to 1.8\%, 
showing that our method
also learns more robust features than standard approaches which ignore
modality information. Even though using the first-pass network alone
underperforms the second-pass network alone by up to 8.3\%, integrating
both in our two-pass system outperforms the solo second-pass network by
up to 2.4\%. This shows that the two-pass system improves not only
efficiency, but accuracy as well. The performance gains from multimodal
regularization and the two-pass system are discussed in detail below.

Our system outperforms all baseline approaches by all metrics
except for the point metric in the object-wise split case. However, we
can see that the chance performance is much higher for the 
point metric than for the 
rectangle metric. This shows that the point metric can overstate
performance, and the rectangle metric is a better indicator of the
accuracy of a grasp detection system.

\begin{figure}[tb]
\begin{center}
\begin{tabular}{m{0.022in} m{0.24in} m{0.8in} m{0.8in} m{0.8in}}
\raisebox{0in}{\rotatebox{90}{\scriptsize{Wide gripper}}} & 
\includegraphics[height=0.5in, width=0.4in]{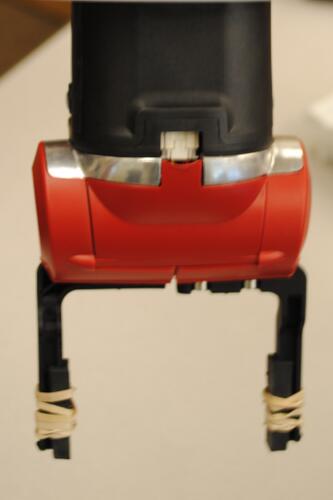} &
\includegraphics[height=0.5in]{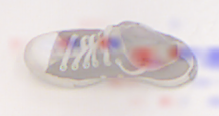} &
\includegraphics[height=0.5in]{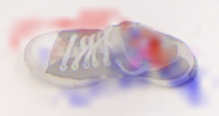} & 
\includegraphics[height=0.5in]{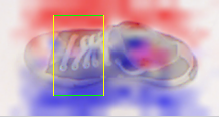} \\
\raisebox{0in}{\rotatebox{90}{\scriptsize{Thin gripper}}} &
\includegraphics[height=0.5in, width=0.4in]{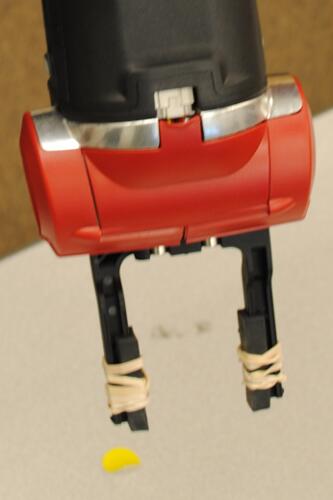} &
\includegraphics[height=0.5in]{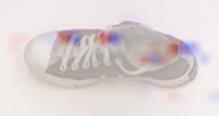} &
\includegraphics[height=0.5in]{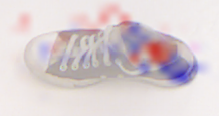} & 
\includegraphics[height=0.5in]{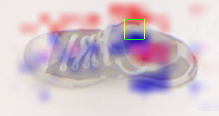}\\
 & & {\vspace*{-0.12in}\begin{center}\footnotesize \hspace{1.5em} 0 degrees\end{center}} & {\vspace*{-0.12in}\begin{center}\footnotesize \hspace{1.25em} 45 degrees\end{center}} & {\vspace*{-0.12in}\begin{center}\footnotesize \hspace{1.25em} 90 degrees\end{center}}
\end{tabular}
\end{center}
\vspace*{-0.2in}
\vspace{\captionReduceTop}
\caption{\textbf{Visualization of grasping scores for different grippers:}
Red indicates maximum score for a grasp with left gripper plane centered
at each point, blue is similar for the right plate. Best-scoring
rectangle shown in green/yellow.}
\vspace{\captionReduceBot}
\label{fig:heatmaps}
\end{figure}

\begin{figure}[t]
\begin{center}
\includegraphics[height=.8in]{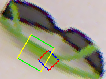}
\includegraphics[width=.8in,angle=90]{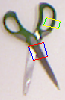}
\includegraphics[height=.8in]{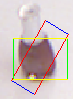}\\
\includegraphics[height=.8in]{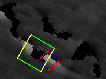}
\includegraphics[width=.8in,angle=90]{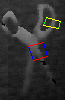}
\includegraphics[height=.8in]{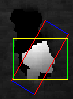}
\end{center}
\vspace{\captionReduceTop}
\vspace{-0.07in}
\caption{\textbf{Improvements from group regularization:} 
Cases where our group regularization approach  produces a viable
grasp (shown in green and yellow), 
while a network trained only with simple \Lone {} regularization does not
(shown in blue and red). Top: RGB image, bottom: depth channel. Green and
blue edges correspond to gripper.}
\vspace*{\captionReduceBot}
\vspace*{\figureReduceBot}
\label{fig:groupBeatsL1}
\end{figure}

\header{Adaptability:}
One important advantage of our detection system is that we can flexibly specify the constraints
of the gripper in our detection system. This is particularly important for
a robot like Baxter, where different objects might require different gripper
settings to grasp. 
We can constrain the detectors to handle this. Figure~\ref{fig:heatmaps} 
shows detection scores
for systems constrained based on two different settings of Baxter's gripper,
one wide and one thin. The implications of these results for other types
of grippers will be discussed in Section~\ref{sec:future}.

\vspace*{\subsectionReduceTop}
\subsection{Multimodal Group Regularization}
\vspace*{\subsectionReduceBot}
Our group regularization term improves detection accuracy over simple \Lone{}
regularization. The improvement is more significant for the object-wise split than for the image-wise split because the group regularization helps the network to avoid overfitting, which will tend to occur more when the learning algorithm is evaluated on unseen objects. 

Figure~\ref{fig:groupBeatsL1}
shows typical cases where a network trained using our 
group regularization
finds a valid grasp, but a network trained with \Lone {} regularization
does not. In these cases, the grasp chosen by the \Lone-regularized network
appears valid for some modalities -- the depth channel for the sunglasses
and nail polish bottle, and the RGB channels for the scissors.
However, when all modalities are considered, the grasp is clearly invalid. The
group-regularized network does a better job of combining information from
all modalities and is more robust to noise and missing data in the depth
channel, as seen in these cases.

\begin{figure}[t]
\vspace*{\figureReduceBot}
\begin{center}
\includegraphics[height=0.8in]{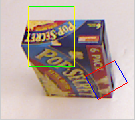}
\includegraphics[height=0.8in]{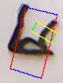}
\includegraphics[height=0.8in]{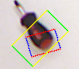}
\includegraphics[height=0.8in]{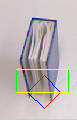}\\
\includegraphics[height=0.8in]{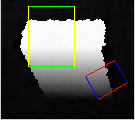}
\includegraphics[height=0.8in]{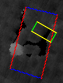}
\includegraphics[height=0.8in]{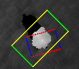}
\includegraphics[height=0.8in]{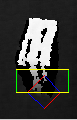}
\end{center}
\vspace*{-0.07in}
\vspace*{\captionReduceTop}
\caption{\textbf{Improvements from two-stage system:} Example cases 
where the two-stage system produces a viable
grasp (shown in green and yellow), 
while the single-stage system does not
(shown in blue and red). Top: RGB image, bottom: depth channel. Green and
blue edges correspond to gripper.}
\vspace*{\captionReduceBot}
\vspace*{\figureReduceBot}
\label{fig:twoPassBeatsOne}
\end{figure}

\vspace*{\subsectionReduceTop}
\subsection{Two-stage Detection System}
\vspace*{\subsectionReduceBot}
Using our two-pass system enhanced both computational performance and
accuracy. The number of rectangles the
full-size network needed to evaluate was reduced by roughly a factor of 1000.
Meanwhile, detection performance increased by up to 2.4\% as
compared to a single pass with the large-size network, even though using
the small network alone significantly underperforms the larger network.
In most cases, the
top 100 rectangles from the first pass contained the top-ranked rectangle from
an exhaustive search using the second-stage network, and thus results were
unaffected.

Figure~\ref{fig:twoPassBeatsOne} shows some cases where the first-stage
network pruned away rectangles corresponding to weak grasps which might
otherwise be chosen by the second-stage network. In these cases, the
grasp chosen by the single-stage system might be feasible for a robotic
gripper, but the rectangle chosen by the two-stage system represents a
grasp which would clearly be successful. 

The two-stage system also significantly increases the computational efficiency
of our detection system. Average inference time for a MATLAB implementation 
of the deep network 
was reduced from 24.6s/image for an exhaustive search using the larger network
to 13.5s/image using the two-stage system.

\begin{figure}[tb]
\begin{center}
\includegraphics[width=0.45\textwidth]{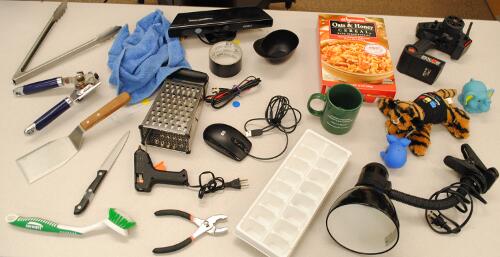}
\end{center}
\caption{\textbf{Robotic experiment objects:}
Several of the objects used in experiments, including challenging
cases such as an oddly-shaped RC car controller, a cloth towel, plush cat, and
white ice cube tray.}
\label{fig:expObj}
\vspace*{\captionReduceBot}
\vspace*{\figureReduceBot}
\end{figure}

\afterpage{
\begin{table*}[t!]
\setlength{\tabcolsep}{4.5pt}
\begin{center}
\captionsetup{justification=centering}
\caption{\textbf{Results for robotic experiments for Baxter,} sorted by object category, for a total of 100 trials. \\ Tr. indicates number of trials, Acc. indicates accuracy
(in terms of success percentage.)}
\label{tbl:expResults}
{\footnotesize
\begin{tabular}{l|r|r||l|r|r||l|r|r||l|r|r||l|r|r} \whline{\tabTopBotLnWd}
\multicolumn{3}{c||}{\textbf{Kitchen tools}} & \multicolumn{3}{c||}{\textbf{Lab tools}}
& \multicolumn{3}{c||}{\textbf{Containers}} & \multicolumn{3}{c||}{\textbf{Toys}} & \multicolumn{3}{c}{\textbf{Others}}\\ \hline
Object & Tr. & Acc. & Object & Tr. & Acc. & Object &Tr. & Acc. & Object & Tr. & Acc. & Object & Tr. & Acc.\\ \hline 
Can opener    & 3 & 100 & Kinect            & 5 & 100 & Colored cereal box   & 3 & 100 & Plastic whale     & 4 &  75 & Electric shaver & 3 & 100 \\
Knife         & 3 & 100 & Wire bundle       & 3 & 100 & White cereal box & 4 &  50 & Plastic elephant  & 4 & 100 & Umbrella        & 4 &  75 \\
Brush         & 3 & 100 & Mouse             & 3 & 100 & Cap-shaped bowl  & 3 & 100 & Plush cat         & 4 &  75 & Desk lamp       & 3 & 100 \\
Tongs         & 3 & 100 & Hot glue gun      & 3 &  67 & Coffee mug       & 3 & 100 & RC controller     & 3 &  67 & Remote control  & 5 & 100 \\
Towel         & 3 & 100 & Quad-rotor        & 4 &  75 & Ice cube tray    & 3 & 100 & XBox controller   & 4 &  50 & Metal bookend   & 3 &  33 \\
Grater        & 3 & 100 & Duct tape roll    & 4 & 100 & Martini glass    & 3 &   0 & Plastic frog      & 3 &  67 & Glove           & 3 & 100 \\ \hline 
\textbf{Average}& & 100 & \textbf{Average}  &   &  90 & \textbf{Average} &   &  75 & \textbf{Average}  &   &  72 & \textbf{Average}&   &  85\\ \hline 
& \\[-0.8em] \hline 
\textbf{Overall} & &  84 \\ \whline{\tabTopBotLnWd} 
\end{tabular}}
\end{center}
\vspace*{\captionReduceBot}
\vspace*{\figureReduceBot}
\end{table*}

\begin{table*}[t!]
\setlength{\tabcolsep}{4.5pt}
\vspace{0.15in}
\begin{center}
\captionsetup{justification=centering}
\caption{\textbf{Results for robotic experiments for PR2,} sorted by object category, for a total of 100 trials. \\ Tr. indicates number of trials, Acc. indicates accuracy
(in terms of success percentage.)}
\label{tbl:pr2Results}
{\footnotesize
\begin{tabular}{l|r|r||l|r|r||l|r|r||l|r|r||l|r|r} \whline{\tabTopBotLnWd}
\multicolumn{3}{c||}{\textbf{Kitchen tools}} & \multicolumn{3}{c||}{\textbf{Lab tools}}
& \multicolumn{3}{c||}{\textbf{Containers}} & \multicolumn{3}{c||}{\textbf{Toys}} & \multicolumn{3}{c}{\textbf{Others}}\\ \hline
Object & Tr. & Acc. & Object & Tr. & Acc. & Object &Tr. & Acc. & Object & Tr. & Acc. & Object & Tr. & Acc.\\ \hline 
Can opener    & 3 & 100 & Kinect            & 5 & 100 & Colored cereal box   & 3 & 100 & Plastic whale     & 4 &  75 & Electric shaver & 3 & 100 \\
Knife         & 3 & 100 & Wire bundle       & 3 & 100 & White cereal box & 4 & 100 & Plastic elephant  & 4 & 100 & Umbrella        & 4 & 100 \\
Brush         & 3 & 100 & Mouse             & 3 & 100 & Cap-shaped bowl  & 3 & 100 & Plush cat         & 4 & 100 & Desk lamp       & 3 & 100 \\
Tongs         & 3 & 100 & Hot glue gun      & 3 &  67 & Coffee mug       & 3 & 100 & RC controller     & 3 &  67 & Remote control  & 5 & 100 \\
Towel         & 3 & 100 & Quad-rotor        & 4 & 100 & Ice cube tray    & 3 & 100 & XBox controller   & 4 &  25 & Metal bookend   & 3 &  67 \\
Grater        & 3 & 100 & Duct tape roll    & 4 & 100 & Martini glass    & 3 &   0 & Plastic frog      & 3 &  67 & Glove           & 3 & 100 \\ \hline 
\textbf{Average}& & 100 & \textbf{Average}  &   &  95 & \textbf{Average} &   &  83 & \textbf{Average}  &   &  72 & \textbf{Average}&   &  95\\ \hline 
& \\[-0.8em] \hline 
\textbf{Overall} & & 89 \\ \whline{\tabTopBotLnWd} 
\end{tabular}}
\end{center}
\end{table*}

\newcommand{\graspFigHt}{1.39in}

\begin{figure*}[t!]
\begin{center}
\includegraphics[height=\graspFigHt]{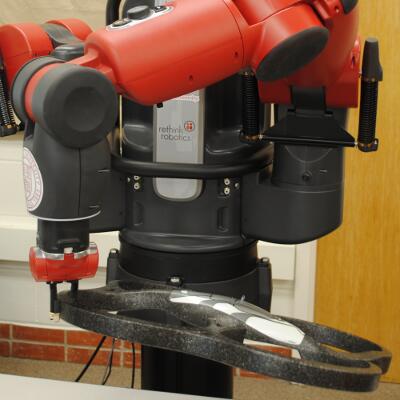}
\includegraphics[height=\graspFigHt]{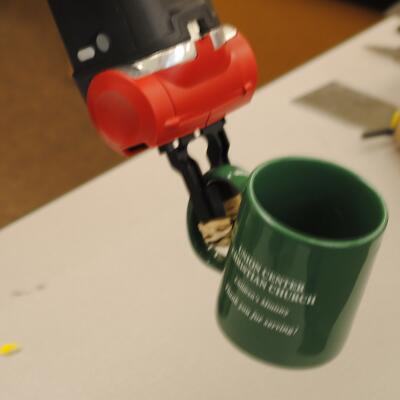}
\includegraphics[height=\graspFigHt]{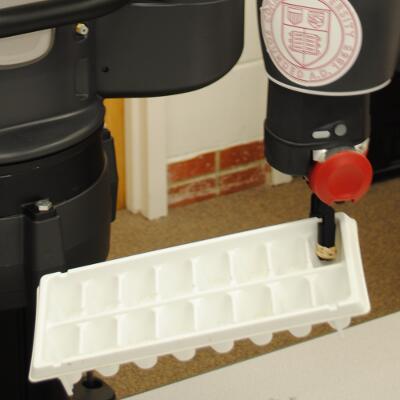} 
\includegraphics[height=\graspFigHt]{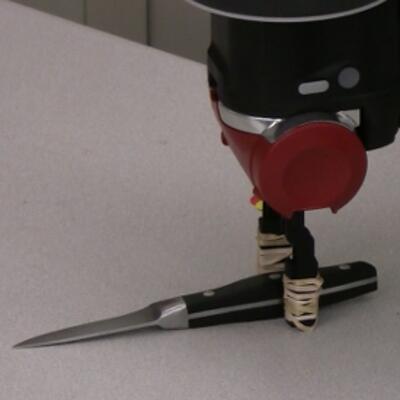}
\includegraphics[height=\graspFigHt]{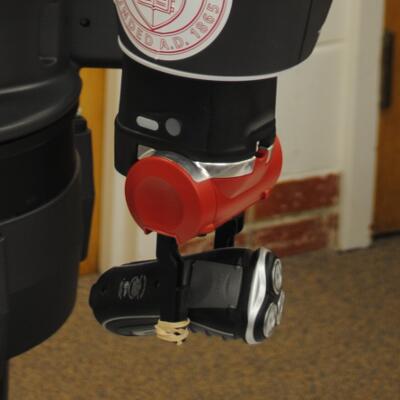} \\ \vspace*{0.05in}
\includegraphics[height=\graspFigHt]{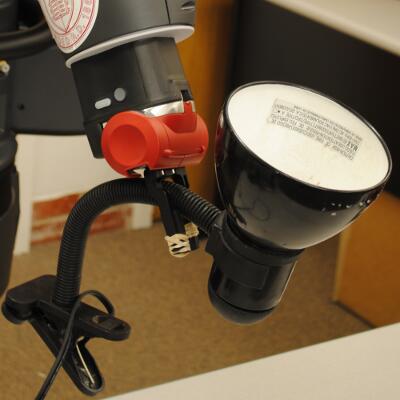}
\includegraphics[height=\graspFigHt]{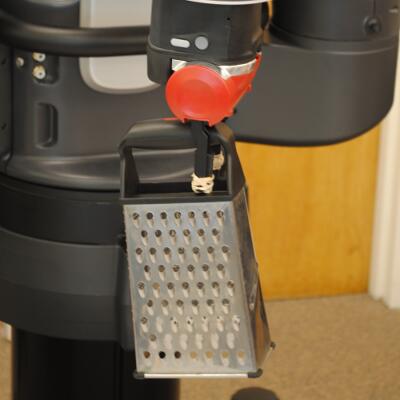}
\includegraphics[height=\graspFigHt]{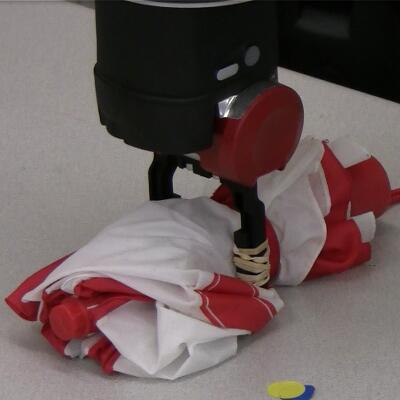}
\includegraphics[height=\graspFigHt]{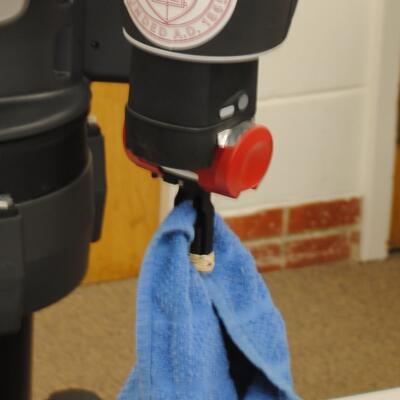}
\includegraphics[height=\graspFigHt]{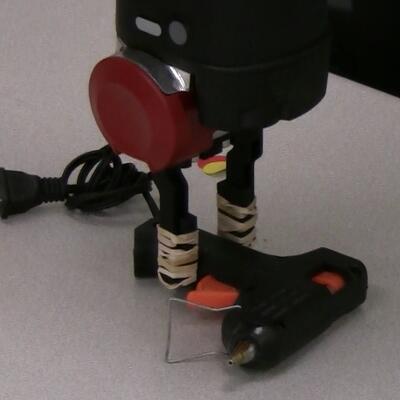} \\ \vspace*{0.05in}
\includegraphics[height=\graspFigHt]{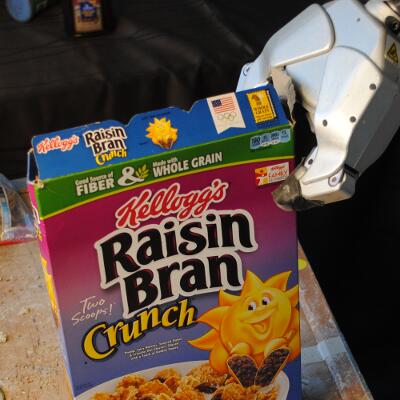}
\includegraphics[height=\graspFigHt]{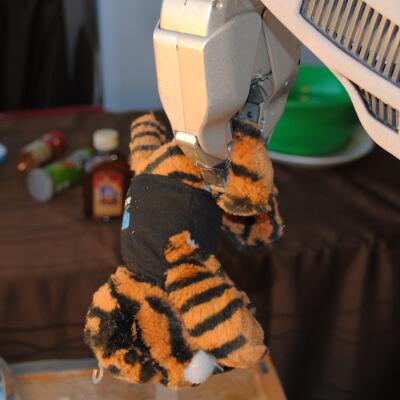}
\includegraphics[height=\graspFigHt]{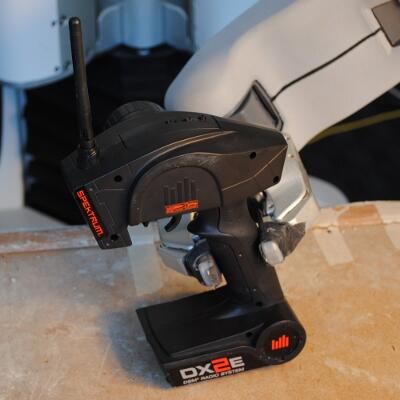}
\includegraphics[height=\graspFigHt]{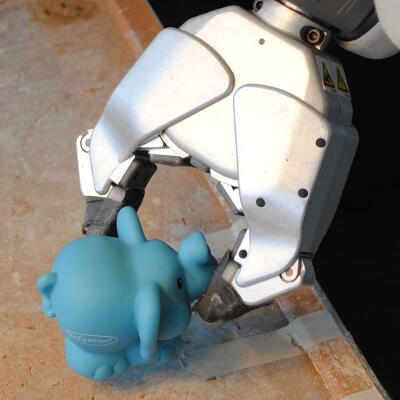}
\includegraphics[height=\graspFigHt]{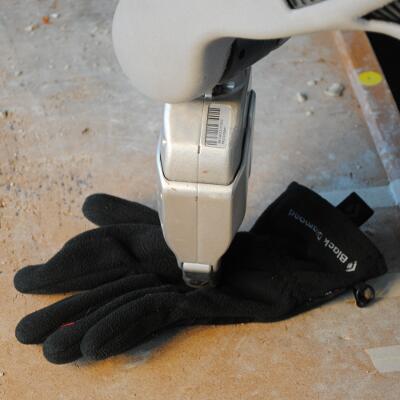}
\end{center}
\caption{\textbf{Robots executing grasps: }
Our robots grasping several objects from the experimental dataset. 
Top row:
Baxter grasping a quad-rotor casing, coffee mug, ice cube tray, knife, and
electric shaver. 
Middle row: Baxter grasping a desk lamp, cheese grater, 
umbrella, cloth towel, and hot glue gun. 
Bottom row: PR2 grasping a plush cat, RC car
controller, cereal box, toy elephant, and glove.}
\label{fig:yogiGrasping}
\end{figure*}
} 


\section{Robotic Experiments\label{sec:baxter}}

In order to evaluate the performance of our algorithms in the real world,
we ran an extensive series of robotic experiments. \newTxt{To explore the
generalizability and effect of the robot on the success rate of our
algorithms, we performed experiments on two different robotic platforms,
a Baxter Research Robot (``Yogi'') and a PR2 (``Kodiak'').}

\header{Baxter:} 
The first platform used is our
Baxter Research Robot, which we call ``Yogi.'' 
Baxter has two arms with seven degrees of
freedom each and a maximum reach of 104 cm, although we used only the
left arm for these experiments. The end-effector for this arm is a two-finger
parallel gripper. We augmented the gripper tips using rubber bands for
additional friction. Baxter's grippers are interchangable,
and we used two settings for these experiments - a ``wide'' setting with an
open width of 8 cm and closed width of 4 cm, and a ``thin''
setting with an open width of 4 cm and a closed width of 0 cm 
(completely closed, gripper tips touching).

To detect grasps, we mounted a Kinect sensor to Yogi's head,
approximately 1.75 m above the ground.
angled downwards at roughly a $75^o$ angle
 towards a table in front of it. The Kinect gives \RGBD images at a resolution
 of 640x480 pixels. We calibrated the 
transformation between the Kinect's and Yogi's coordinate frames by marking
four points corresponding to a set of 3D axes, and obtaining the
coordinates of these points in both Kinect's and Yogi's frames.

All control for Baxter was done by specifying an end-effector 
position and orientation,
and using the inverse kinematics provided with Baxter to determine a set of 
joint angles for this pose. Baxter's built-in control systems were used
to drive the arm to these new joint angles.  

\header{PR2:}
\newTxt{
Our second platform was our PR2 robot, ``Kodiak.'' Similar to Baxter, PR2 has
two 7-DoF arms with approximately 1 m reach, and we used only the left for
these experiments. PR2's grippers open to a width of 8 cm, and are
capable of closing completely from that span, so we did not need to use two
settings as with Baxter. We augmented PR2's gripper friction with gaffer
tape on the fingertips.

For the experiments on PR2, we used the Kinect already mounted to Kodiak's head,
and used ROS's built-in functionality to obtain 3D locations from that Kinect
and transform these to Kodiak's body frame for manipulation. Control was
performed using the ee\_cart stiffness controller \cite{bakebot} 
with trajectories provided by our own custom MATLAB code. 
} 

\header{Experimental Setup:}
For each experiment, we placed a single object within a 25 cm x 25 cm square on 
the table, approximately 1.2 m below the mounting point of the Kinect. This
square was chosen to be
well-contained within each robot's workspace, allowing objects to be reached from
most approach vectors. Object positions and orientations were varied between trials,
although objects were always placed in configurations in which at least one
viable grasp was visible and accessible to the robot. 

When using Baxter, due to the limited stroke (span from open to closed) of its
gripper, 
we pre-selected one of the two gripper
settings discussed above for each object. We constrained the search space
as illustrated in Fig.~\ref{fig:heatmaps} to find grasps for that particular
setting.

To detect grasps, we first took an \RGBD image from the Kinect with no objects
in the scene as a background image. The depth channel of this image was used to
segment objects from the scene, and to correct for the slant of the Kinect.
Once an object was segmented, we used our algorithm, as described above, to
obtain a single best-ranked grasping rectangle. 

\newTxt{The search space for the
first-pass network progressed in 15-degree increments from 15 to 180 degrees
(angles larger than 180 being mirror-images of grasps already tested), searching
over 10-pixel increments across the image for the X and Y coordinates of the 
upper-left corner of the rectangle. For the thin gripper setting, rectangle
widths and heights from 10 to 40 pixels in 10-pixel increments were searched,
while for the thick setting these ranged from 40 pixels to 100 pixels in 
20-pixel increments. In both cases, rectangles taller than they were wide were
ignored. Once a single best-scoring grasp was detected, we translated it to a
robotic grasp consisting of a grasping point and an approach vector using the
rectangle's parameters and the surface normal at the rectangle's center as
described above.}

To execute the grasp, we first positioned the gripper at a location 10 cm back
from the grasping point along the approach vector. The gripper was oriented to
the approach vector, and rotated around it based on the orientation of the
detected grasping rectangle.

Since Baxter's arms are highly compliant, slight imprecisions in end-effector
positioning are to be expected -- we found that errors of up to 2 cm were 
typical.
Thus, we implemented a visual servoing system using its hand camera, which
provides RGB images at a resolution of 320x200 pixels. We used color 
segmentation to separate the object from the background, and used its lateral
position in image space to drive Yogi's end-effector to center the object.
We did not implement visual servoing for PR2 because its gripper positioning
was found to be precise to within 0.5 cm.

After visual servoing was completed, we drove the gripper 14 cm forwards from its
current position along the approach vector, so that the grasping point was
well-contained within it. We then closed the gripper, grasping the
object, and moved it 30 cm upwards.
A grasp was
determined to be successful if it was sufficient to lift the object and hold it
for one second.

\header{Objects to be Grasped:}
For our robotic experiments, we collected a diverse set of 35 objects within a size of
.3 m x .3 m x .3 m and weighing at most 2.5 kg (although most were
less than 1 kg) from our offices, homes, and lab. 
Many of them are shown in Fig.~\ref{fig:expObj}. 
Most of these objects were not present in the
training dataset, and thus were completely new to the grasp detection algorithm.

Due to the physical limitations of the robots' grippers, 
we found that five of these objects were not graspable 
even when given a hand-chosen grasp.
The small pair of pliers was too low to
the table to grip properly. 
The spray paint can was too smooth for the gripper to get enough
friction to lift it. The weight of the hammer was too imbalanced,
causing the hammer to rotate and slip out of the gripper when grasped.
Similar problems were encountered with the bicycle U-lock.
The bevel spatula's handle was
too close to the thin-set size of Baxter's gripper, so that we could not
position it precisely enough to grasp it reliably. We did not consider these 
objects for purposes of our experimental results, since our focus was on
evaluating the performance of our grasp detection algorithm.

\header{Results:}
Table~\ref{tbl:expResults} shows the results of our robotic experiments on
Baxter for
the remaining 30 objects, a total
of 100 trials. Using our algorithm, Yogi was able
to successfully execute a grasp in 84\% of the trials.
Figure~\ref{fig:yogiGrasping} shows Yogi executing
several of these grasps. In 8\% of the trials,
our algorithm detected a valid grasp which was not executed correctly by
Yogi. Thus, we were able to successfully detect a good grasp in
92\% of the trials. Video of some of these trials is available at
\texttt{\url{http://pr.cs.cornell.edu/deepgrasping}}.

\newTxt{
PR2 yielded a higher success rate as seen in Table~\ref{tbl:pr2Results},
succeeding in 89\% of trials. This is 
largely due to the much wider span of PR2's gripper from open to closed and
its ability to fully close from its widest position, as well as PR2's ability
to apply a larger gripping force. Some specific instances where PR2 and
Baxter's performance differed are discussed below.
} 

For comparison purposes, we ran a small set of control experiments for
16 of the objects in the dataset. The control algorithm simply returned
a fixed-size rectangle centered at the object's center of mass, as
determined by depth segmentation from the background. The rectangle was
aligned so that the gripper plates ran parallel to the object's
principal axis. This algorithm was only successful in 31\% of cases,
significantly underperforming our system.

On Baxter, our algorithm sometimes detected a grasp which was not realizable
by the
current setting of its gripper, but might be executable by others. For
example, our algorithm detected grasps across the leg of the plush cat, 
and the region between the handle and body of the umbrella,
both too thin for the wide setting of Baxter's
gripper to grasp since it has a minimum span of 4 cm. \newTxt{Since PR2's 
gripper can close completely from any position, it did not encounter these
issues and thus achieved a 100\% success rate for both these objects.}

\newTxt{
The XBox controller proved to be a very difficult object for either robot
to grasp. From a top-down angle, there is only a small space of
viable grasps with a span of less than 8 cm, but many which have either a
slightly larger span (making them non-realizable by either gripper), or 
are subtly non-viable (e.g. grasps across the two ``handles,'' which tend
to slip off.)  All viable grasps are very near to the 8 cm span of both
grippers, meaning that even slight imprecision in positioning can lead to
failure. Due to this, Baxter achieved a higher success rate for the XBox
controller thanks to visual servoing, succeeding in 50\% of cases as 
compared to the 25\% success rate for PR2.
} 

Our algorithm was able to consistently detect and execute valid grasps for a
red cereal box, but had some failures on a white and yellow one. This is because
the background for all objects in the dataset is white, leading the algorithm
to learn features relating white areas at the edges of the gripper region to
graspable cases. However, it was able to detect and execute correct grasps for
an all-white ice cube tray, and so does not fail for all white objects. This
could be remedied by extending the dataset to include cases with different
background colors.
\newTxt{Interestingly, even though the parameters of grasps detected
for the white box were similar for PR2 and Baxter, PR2 was able to succeed in
every case while Baxter succeeded only half the time. This is because PR2's
increased gripper strength allowed it to execute grasps across corners of the
box, crushing it slightly in the process.}

Other failures were due to the limitations of the Kinect sensor. We were never
able to properly grasp the martini glass because its glossy finish prevented Kinect
from returning any depth estimates for it. Even if a valid grasp were detected using
color information only, there was no way to infer a proper grasping position without
depth information. Grasps for the metal bookend failed for similar reasons, but it
was not as glossy as the martini glass, and gave enough returns for some to succeed.

However, our algorithm also had many noteworthy successes. It was able to
consistently detect and execute grasps for a crumpled cloth towel, a
complex and irregular case which bore little resemblance to any object in
the dataset. It was also able to find and grasp the rims of objects such as
the plastic baseball cap and coffee mug, cases where there is little visual
distinction between the rim and body of the object. These objects underscore the
importance of the depth channel for robotic grasping, as none of these grasps
would be detectable without depth information.

\newTxt{
Our algorithm was also able to successfully detect and execute many grasps for
which the approach vector was non-vertical. The grasps shown for the
coffee mug, desk lamp, cereal box, RC car controller, and toy elephant
shown in Fig.~\ref{fig:yogiGrasping} 
were all executed by aligning the gripper to such an approach vector.
Indeed, many of these grasps may have failed had the gripper been aligned
vertically. This shows that our algorithm is not restricted to detecting
top-down grasps, but rather encodes a more general notion of graspability which
can be applied to grasps from many angles, albeit within the constraints of
visibility from a single-view perspective.
}

While a few failures
occurred, our algorithm still achieved a high rate of accuracy for 
other oddly-shaped
objects such as the quad-rotor casing, RC car controller, and glue gun.
For objects with clearly defined handles, such as the cheese grater, kitchen
tongs, can opener, and knife, our algorithm was able to detect and execute
successful grasps in every trial, showing that there is a wide range of
objects which it can grasp extremely consistently.

\vspace*{\sectionReduceTop}
\section{Discussion and Future Work\label{sec:future}}
\vspace*{\sectionReduceBot}
\newTxt{
Our algorithm focuses on the problem of grasp detection for a two-fingered
parallel-plate style gripper. It would be directly applicable to other
grippers with fixed configurations, simply requiring new training data
labeled with grasps for the gripper in question. Our system would allow
even the basic features used for grasp detection to adapt to the gripper.
This might be useful in cases such as jamming grippers \cite{YunJamming}, or 
two-fingered grippers with differently-shaped contact surfaces, which
might require different features to determine a graspable area.

Our detection algorithm does not directly address the problem of 3D orientation
of the gripper -- this orientation is determined only after an optimal
rectangle has been detected, orienting the grasp based on the object's surface
normals. However, just as our approach here considers aligns a 2D feature 
window to the gripper, an extension of this work might align a 3D window -- 
using voxels, rather than pixels, as its basic unit of representation for
input features to the network. This would allow the system to search across
the full 6-DoF 3D pose of the gripper, while still leveraging the power of
feature learning. 

Our system gives only a gripper pose as output, but multi-fingered 
reconfigurable hands also require a configuration of the fingers in order
to grasp an object. In this case,
our algorithm could be used as a heuristic to find one or more locations 
likely to be graspable (similar to the first pass in our two-pass system),
greatly reducing the search space needed to find
an optimal gripper configuration.

Our algorithm also depends only on local features to determine grasping
locations. However, many household objects may have some areas which are
strongly preferable to grasp over others - for example, a knife might be
graspable by the blade, or a hot glue gun by the barrel, but both should
actually be grasped by their respective handles. Since these regions are
more likely to be labeled as graspable in the data, our system already
weakly encodes this, but some may not be readily distinguishable
using only local information. Adding a term modeling the probability of each
region of the image being a semantically-appropriate area to grasp the
object would allow us to incorporate this information. This term could be
computed once for the entire image, then added to each local detection score,
keeping detection efficient.

In this work, our visual-servoing algorithm was purely heuristic, simply
attempting to center the segmented object underneath the hand camera.
However, in future work, a similar feature-learning approach might be 
applied to hand camera images of graspable and non-graspable regions,
improving the visual servoing system's ability to fine-tune gripper position
to ensure a good grasp.
} 

Many robotics problems require the use of perceptual information, but can
be difficult and time-consuming to engineer good features for, particularly
when using \RGBD data. 
In future work, our approach 
could be extended to a wide range of such problems. Our system could easily
be applied to other detection problems such as object detection
 or obstacle detection. However, it could also be adapted to
other similar problems, such as object tracking and visual servoing.

Multimodal data has become extremely important for robotics, due both
to the advent of new sensors such as the Kinect and the application of
robots to more challenging tasks which require multiple modalities of
information to perform well. However, it can be very difficult to 
design features
which do a good job of integrating many modalities. While
our work focuses on color, depth, and surface normals as input modes,
our structured multimodal regularization algorithm might also be applied to
others. 
This approach could improve performance while allowing roboticists to focus on
other engineering challenges.

\vspace*{\sectionReduceTop}
\section{Conclusions\label{sec:conclusion}}
\vspace*{\sectionReduceBot}


We presented a system for detecting robotic grasps from \RGBD data
using a deep learning approach. 
Our method has several
advantages over current state-of-the-art methods. First, using deep learning
allows us to avoid hand-engineering features, learning them instead. 
Second, our results show that deep learning methods significantly 
outperform even well-designed hand-engineered features from previous work.

We also presented a novel feature learning algorithm for multimodal data based 
on group regularization. In extensive experiments, we demonstrated that this
algorithm produces better features for robotic grasp detection than existing
deep learning approaches to multimodal data. Our experiments and 
results, both offline and on real robotic platforms,
show that our two-stage deep learning system with group regularization
is capable of robustly detecting grasps for a wide range of objects, even
those previously unseen by the system.

\vspace*{\sectionReduceTop}
\section*{Acknowledgements}
\vspace*{\sectionReduceBot}
We would like to thank Yun Jiang and Marcus Lim for useful discussions and help
with baseline experiments.
 This research was funded in part by ARO award
W911NF-12-1-0267, Microsoft Faculty Fellowship and 
NSF CAREER Award (Saxena), and Google Faculty Research Award (Lee).

{ 
 \bibliographystyle{abbrvnat}
\bibliography{bibliography}
}

\end{document}